\documentclass[twoside,11pt]{article}

\usepackage{jmlr2e}
\usepackage{pslatex}

\usepackage{amsmath}
\usepackage{cases}
\usepackage{bigints}
\usepackage{url}

\renewcommand{\leq}{\leqslant}
\renewcommand{\geq}{\geqslant}
\renewcommand{\phi}{\varphi}
\renewcommand{\hat}{\widehat}

\newcommand{\bu}{\boldsymbol{u}}
\newcommand{\bv}{\boldsymbol{v}}

\newcommand{\bx}{\boldsymbol{x}}

\newcommand{\bphi}{\boldsymbol{\phi}}

\newcommand{\cM}{\mathcal{M}}
\newcommand{\cN}{\mathcal{N}}

\newcommand{\cU}{\mathcal{U}}

\newcommand{\cX}{\mathcal{X}}

\renewcommand{\d}{\mbox{d}}
\newcommand{\dd}{\mbox{\rm d}}

\newcommand{\E}{\mathbb{E}}

\newcommand{\N}{\mathbb{N}}

\newcommand{\Prob}{\mathbb{P}}
\newcommand{\R}{\mathbb{R}}

\newcommand{\wh}{\widehat}

\renewcommand{\epsilon}{\varepsilon}

\newcommand{\norm}[1][\cdot]{\ensuremath{\left\Arrowvert #1 \right\Arrowvert}}
\providecommand{\Norm}[1]{\left\lVert#1\right\rVert}

\newcommand{\indic}[1]{\mathbb{I}_{\{#1\}}}

\newcommand{\eqdef}{\triangleq}

\newcommand{\KL}{\mathcal{K}}
\newcommand{\Var}{\textrm{Var}}

\renewcommand{\tilde}{\widetilde}

\newlength{\minipagewidth}
\newcommand{\bookbox}[1]{
\par\medskip\noindent
\framebox[\textwidth]{
\begin{minipage}{\minipagewidth}
{#1}
\end{minipage} } \par\medskip }

\newenvironment{proofref}[1]{\par\noindent{\bf Proof (of~#1)\ }}{\hfill\BlackBox\\[2mm]}

\newenvironment{proofsketchref}[1]{\par\noindent{\bf Proof sketch (of~#1)\ }}{\hfill\BlackBox\\[2mm]}

\jmlrheading{14}{2013}{729-769}{3/12; Revised 11/12}{3/13}{S\'{e}bastien Gerchinovitz}

\ShortHeadings{Sparsity Regret Bounds for Individual Sequences}{Gerchinovitz}
\firstpageno{729}

\begin{document}

\title{Sparsity Regret Bounds for Individual Sequences \\ in Online Linear Regression\thanks{A shorter version appeared in the proceedings of COLT 2011 (see \citealt{Ger-11colt-SparsityRegretBounds}).}}

\author{\name S\'{e}bastien Gerchinovitz \email sebastien.gerchinovitz@ens.fr \\
       \addr {\'E}cole Normale Sup{\'e}rieure\thanks{This research was
    carried out within the INRIA project CLASSIC hosted by {\'E}cole
    Normale Sup{\'e}rieure and CNRS.}\\
    45 rue d'Ulm \\ Paris, FRANCE}

\editor{Nicol\`{o} Cesa-Bianchi}

\maketitle

\begin{abstract}
We consider the problem of online linear regression on arbitrary deterministic sequences when the ambient dimension $d$ can be much larger than the number of time rounds $T$. We introduce the notion of \emph{sparsity regret bound}, which is a deterministic online counterpart of recent risk bounds derived in the stochastic setting under a sparsity scenario. We prove such regret bounds for an online-learning algorithm called SeqSEW and based on exponential weighting and data-driven truncation. In a second part we apply a parameter-free version of this algorithm to the stochastic setting (regression model with random design). This yields risk bounds of the same flavor as in \citet{DaTsy10MirrorAveraging} but which solve two questions left open therein. In particular our risk bounds are adaptive (up to a logarithmic factor) to the unknown variance of the noise if the latter is Gaussian. We also address the regression model with fixed design.
\end{abstract}


\section{Introduction}
\label{sec:jmlr-intro}

Sparsity has been extensively studied in the stochastic setting over the past decade. This notion is key to address statistical problems that are high-dimensional, that is, where the number of unknown parameters is of the same order or even much larger than the number of observations. This is the case in many contemporary applications such as computational biology (e.g., analysis of DNA sequences), collaborative filtering (e.g., Netflix, Amazon), satellite and hyperspectral imaging, and high-dimensional econometrics (e.g., cross-country growth regression problems).

A key message about sparsity is that, although high-dimensional statistical inference is impossible in general (i.e., without further assumptions), it becomes statistically feasible if among the many unknown parameters, only few of them are non-zero. Such a situation is called a \emph{sparsity scenario} and has been the focus of many theoretical, computational, and practical works over the past decade in the stochastic setting. On the theoretical side, most sparsity-related risk bounds take the form of the so-called \emph{sparsity oracle inequalities}, that is, risk bounds expressed in terms of the number of non-zero coordinates of the oracle vector. As of now, such theoretical guarantees have only been proved under stochastic assumptions.\footnote{One could object that most high-probability risk bounds derived for $\ell^1$-regularization methods are in fact deterministic inequalities that hold true whenever the noise vector $\epsilon$ belong to some set $S$ (see, e.g., \citealt{BiRiTsy-09-LassoDantzig}). However, the fact that $\epsilon \in S$ with high-probability is only guaranteed via concentration arguments, so it is a consequence of the underlying statistical assumptions.}

In this paper we address the prediction possibilities under a sparsity scenario in both deterministic and stochastic settings. We first prove that theoretical guarantees similar to sparsity oracle inequalities can be obtained in a deterministic online setting, namely, online linear regression on individual sequences. The newly obtained deterministic prediction guarantees are called \emph{sparsity regret bounds}. We prove such bounds for an online-learning algorithm which, in its most sophisticated version, is fully automatic in the sense that no preliminary knowledge is needed for the choice of its tuning parameters. In the second part of this paper, we apply our sparsity regret bounds---of deterministic nature---to the stochastic setting (regression model with random design). One of our key results is that, thanks to our online tuning techniques, these deterministic bounds imply sparsity oracle inequalities that are adaptive to the unknown variance of the noise (up to logarithmic factors) when the latter is Gaussian. In particular, this solves an open question raised by \citet{DaTsy10MirrorAveraging}.

In the next paragraphs, we introduce our main setting and motivate the notion of sparsity regret bound from an online-learning viewpoint. We then detail our main contributions with respect to the statistical literature and the machine-learning literature.

\subsection{Introduction of a Deterministic Counterpart of Sparsity Oracle Inequalities}

We consider the problem of online linear regression on arbitrary deterministic sequences. A forecaster has to predict in a sequential fashion the values $y_t \in \R$ of an unknown sequence of observations given some input data $x_t \in \cX$ and some base forecasters $\phi_j: \cX \to \R$, $1 \leq j \leq d$, on the basis of which he outputs a prediction $\widehat{y}_t \in \R$. The quality of the predictions is assessed by the square loss. The goal of the forecaster is to predict almost as well as the best linear forecaster $\bu \cdot \bphi \eqdef \sum_{j=1}^d u_j \phi_j$, where $\bu \in \R^d$, that is, to satisfy, uniformly over all individual sequences $(x_t,y_t)_{1 \leq t \leq T}$, a regret bound of the form
\[
\sum_{t=1}^T \bigl(y_t - \widehat{y}_t\bigr)^2 \leq \inf_{\bu \in \R^d} \left\{\sum_{t=1}^T \bigl(y_t - \bu \cdot \bphi(x_t)\bigr)^2 + \Delta_{T,d}(\bu)\right\}
\]
for some regret term $\Delta_{T,d}(\bu)$ that should be as small as possible and, in particular, sublinear in $T$. (For the sake of introduction, we omit the dependencies of $\Delta_{T,d}(\bu)$ on the amplitudes $\max_{1 \leq t \leq T} |y_t|$ and $\max_{1 \leq t \leq T} \max_{1 \leq j \leq d} |\phi_j(x_t)|$.)

In this setting the version of the sequential ridge regression forecaster studied by \citet{AzWa01RelativeLossBounds} and \citet{Vo01CompetitiveOnline} can be tuned to have a regret $\Delta_{T,d}(\bu)$ of order at most $d \ln\bigl(T \norm[\bu]_2^2\bigr)$. When the ambient dimension $d$ is much larger than the number of time rounds $T$, the latter regret bound may unfortunately be larger than $T$ and is thus somehow trivial. Since the regret bound $d \ln T$ is optimal in a certain sense (see, e.g., the lower bound of \citealt[Theorem~2]{Vo01CompetitiveOnline}), additional assumptions are needed to get interesting theoretical guarantees.

A natural assumption, which has already been extensively studied in the stochastic setting, is that there is a sparse vector $\bu^*$ (i.e., with $s \ll T/(\ln T)$ non-zero coefficients) such that the linear combination $\bu^* \cdot \bphi$ has a small cumulative square loss. If the forecaster knew in advance the support $J(\bu^*) \eqdef \{j: u^*_j \neq 0\}$ of $\bu^*$, he could apply the same forecaster as above but only to the $s$-dimensional linear subspace $\bigl\{\bu \in \R^d: \forall j \notin J(\bu^*), u_j = 0 \bigr\}$. The regret bound of this ``oracle'' would be roughly of order $s \ln T$ and thus sublinear in $T$. Under this sparsity scenario, a sublinear regret thus seems possible, though, of course, the aforementioned regret bound $s \ln T$ can only be used as an ideal benchmark (since the support of $\bu^*$ is unknown).

In this paper, we prove that a regret bound proportional to $s$ is achievable (up to logarithmic factors). In Corollary~\ref{cor:chapSparse-SRB-known-optimalbound} and its refinements (Corollary~\ref{cor:chapSparse-SRB-unknownBy-tauOptimal} and Theorem~\ref{thm:chapSparse-SRB-unknownByBphi}), we indeed derive regret bounds of the form
\begin{equation}
\label{eqn:chapSparse-bounds-shape}
\sum_{t=1}^T (y_t - \widehat{y}_t)^2 \leq \inf_{\bu \in \R^d} \left\{\sum_{t=1}^T \bigl(y_t - \bu \cdot \bphi(x_t)\bigr)^2 + \bigl(\norm[\bu]_0 +1\bigr) \, g_{T,d}\bigl(\norm[\bu]_1, \norm[\bphi]_{\infty}\bigr)\!\right\}~,
\end{equation}
where $\norm[\bu]_0$ denotes the number of non-zero coordinates of $\bu$ and where $g$ grows at most logarithmically in $T$, $d$, $\norm[\bu]_1 \eqdef \sum_{j=1}^d |u_j|$, and $\norm[\bphi]_{\infty} \eqdef \sup_{x \in \cX} \max_{1 \leq j \leq d} |\phi_j(x)|$. We call regret bounds of the above form \emph{sparsity regret bounds}.

This work is in connection with several papers that belong either to the statistical or to the machine-learning literature. Next we discuss these papers and some related references.

\subsection{Related Works in the Stochastic Setting}

The above regret bound~(\ref{eqn:chapSparse-bounds-shape}) can be seen as a deterministic online counterpart of the so-called \emph{sparsity oracle inequalities} introduced in the stochastic setting in the past decade. The latter are risk bounds expressed in terms of the number of non-zero coordinates of the oracle vector---see \eqref{eqn:jmlr-soi-shape} below. More formally, consider the regression model with random of fixed design. The forecaster observes independent random pairs $(X_1,Y_1),\ldots,(X_T,Y_T) \in \cX \times \R$ given by
\[
Y_t = f(X_t) + \epsilon_t~, \quad 1 \leq t \leq T~,
\]
where the $X_t \in \cX$ are either {i.i.d} random variables (random design) or fixed elements (fixed design), denoted in both cases by capital letters in this paragraph, and where the $\epsilon_t$ are {i.i.d.} square-integrable real random variables with zero mean (conditionally on the $X_t$ if the design is random). The goal of the forecaster is to construct an estimator $\hat{f}_T:\cX \to \R$ of the unknown regression function $f:\cX \to \R$ based on the sample $(X_t,Y_t)_{1 \leq t \leq T}$. Depending on the nature of the design, the performance of $\hat{f}_T$ is measured through its risk $R\bigl(\hat{f}_T\bigr)$:
\vspace{-0.2cm}
\begin{numcases}{R\bigl(\hat{f}_T\bigr) \eqdef}
\nonumber
\int_{\cX} \left(f(x) - \hat{f}_T(x) \right)^2 P^X(\dd x) & (random design) \\
\nonumber
\frac{1}{T} \sum_{t=1}^T \bigl(f(X_t) - \hat{f}_T(X_t)\bigr)^2 & (fixed design),
\end{numcases}
where $P^X$ denotes the common distribution of the $X_t$ if the design is random. With the above notations, and given a dictionary $\bphi = (\phi_1,\cdots,\phi_d)$ of base forecasters $\phi_j:\cX \to \R$ as previously, typical examples of \emph{sparsity oracle inequalities} take approximately the form
\begin{equation}
\label{eqn:jmlr-soi-shape}
R\bigl(\hat{f}_T\bigr) \leq C \inf_{\bu \in \R^d} \left\{R\bigl(\bu \cdot \bphi \bigr) + \frac{\norm[\bu]_0 \ln d + 1}{T}\right\}
\end{equation}
in expectation or with high probability, for some constant $C \geq 1$. Thus, sparsity oracle inequalities are risk bounds involving a trade-off between the risk $R(\bu \cdot \bphi)$ and the number of non-zero coordinates $\norm[\bu]_0$ of any comparison vector $\bu \in \R^d$. In particular, they indicate that $\hat{f}_T$ has a small risk under a sparsity scenario, that is, if $f$ is well approximated by a sparse linear combination $\bu^* \cdot \bphi$ of the base forecasters $\phi_j$, $1 \leq j \leq d$.

Sparsity oracle inequalities were first derived by \citet{BiMa-01-GaussianMS} via $\ell^0$-regularization methods (through model-selection arguments). Later works in this direction include, among many other papers, those of \citet{BiMa-07-MinimalPenalties}, \citet{AbBeDoJo-06-AdaptingUnknownSparsity},  and \citet{BTW-07-AggregationGaussianRegression} in the regression model with fixed design and that of \citet{BTW-04-AggregationRegressionLearning} in the random design case.

More recently, a large body of research has been dedicated to the analysis of $\ell^1$-regularization methods, which are convex and thus computationally tractable variants of $\ell^0$-regularization methods. A celebrated example is the Lasso estimator introduced by \citet{Tibshirani96Lasso} and \citet{DoJo-94-WaveletShrinkage}. Under some assumptions on the design matrix,\footnote{Despite their computational efficiency, the aforementioned $\ell^1$-regularized methods still suffer from a drawback: their $\ell^0$-oracle properties hold under rather restrictive assumptions on the design; namely, that the $\phi_j$ should be nearly orthogonal (see the detailed discussion in \citealt{vdGBu-09-ConditionsLasso}).} such methods have been proved to satisfy sparsity oracle inequalities of the form~\eqref{eqn:jmlr-soi-shape} (with $C=1$ in the recent paper by \citealt{KoLoTsy-11-MatrixCompletion}). A list of few references---but far from being comprehensive---includes the works of \citet{BTW-07-SOILasso}, \citet{CaTa-07-Dantzig}, \citet{vdG-08-glimLasso}, \citet{BiRiTsy-09-LassoDantzig}, \citet{Kol-09-SparseRecoveryEntropy}, \citet{Kol-09-SparsityPenalizationLp},\citet{HevdG-11-AdaptiveLasso}, \citet{KoLoTsy-11-MatrixCompletion} and \citet{LoPovdGTsy-11-GroupSparsity}. We refer the reader to the monograph by \citet{BuvdG-11-HighDim} for a detailed account on $\ell^1$-regularization.

A third line of research recently focused on procedures based on exponential weighting. Such methods were proved to satisfy sharp sparsity oracle inequalities (i.e., with leading constant $C=1$), either in the regression model with fixed design \citep{DaTsy07SEW,DaTsy08SEW,RiTsy-10-ExpScreening,AlLo-10-PACBSparseRegression} or in the regression model with random design \citep{DaTsy10MirrorAveraging,AlLo-10-PACBSparseRegression}. These papers show that a trade-off can be reached between strong theoretical guarantees (as with $\ell^0$-regularization) and computational efficiency (as with $\ell^1$-regularization). They indeed propose aggregation algorithms which satisfy sparsity oracle inequalities under almost no assumption on the base forecasters $(\phi_j)_{j}$, and which can be approximated numerically at a reasonable computational cost for large values of the ambient dimension~$d$.

Our online-learning algorithm SeqSEW is inspired from a statistical method of \citet{DaTsy08SEW,DaTsy10MirrorAveraging}.
Following the same lines as in \citet{DaTsy10MonteCarlo}, it is possible to slightly adapt the statement of our algorithm to make it computationally tractable by means of Langevin Monte-Carlo approximation---without affecting its statistical properties. The technical details are however omitted in this paper, which only focuses on the theoretical guarantees of the algorithm SeqSEW.

\subsection{Previous Works on Sparsity in the Framework of Individual Sequences}

To the best of our knowledge, Corollary~\ref{cor:chapSparse-SRB-known-optimalbound} and its refinements (Corollary~\ref{cor:chapSparse-SRB-unknownBy-tauOptimal} and Theorem~\ref{thm:chapSparse-SRB-unknownByBphi}) provide the first examples of sparsity regret bounds in the sense of~(\ref{eqn:chapSparse-bounds-shape}). To comment on the optimality of such regret bounds and compare them to related results in the framework of individual sequences, note that (\ref{eqn:chapSparse-bounds-shape}) can be rewritten in the equivalent form:

For all $s \in \N$ and all $U > 0$,
\[
\sum_{t=1}^T (y_t - \widehat{y}_t)^2 -  \inf_{\substack{\norm[\bu]_0 \leq s \\ \norm[\bu]_1 \leq U}} \, \sum_{t=1}^T \bigl(y_t - \bu \cdot \bphi(x_t)\bigr)^2 \leq \bigl(s + 1\bigr) \, g_{T,d}\bigl(U, \norm[\bphi]_{\infty}\bigr)~,
\]
where $g$ grows at most logarithmically in $T$, $d$, $U$, and $\norm[\bphi]_{\infty}$. When $s \ll T$, this upper bound matches (up to logarithmic factors) the lower bound of order $s \ln T$ that follows in a straightforward manner from Theorem~2 of \citet{Vo01CompetitiveOnline}. Indeed, if $s \ll T$, $\cX = \R^d$, and $\phi_j(x) = x_j$, then for any forecaster, there is an individual sequence $(x_t,y_t)_{1 \leq t \leq T}$ such that the regret of this forecaster on $\bigl\{\bu \in \R^d: \norm[\bu]_0 \leq s~\textrm{and}~\norm[\bu]_1 \leq d\bigr\}$ is bounded from below by a quantity of order $s \ln T$. Therefore, up to logarithmic factors, any algorithm satisfying a sparsity regret bound of the form~(\ref{eqn:chapSparse-bounds-shape}) is minimax optimal on intersections of $\ell^0$-balls (of radii $s \ll T$) and $\ell^1$-balls. This is in particular the case for our algorithm SeqSEW, but this contrasts with related works discussed below.

Recent works in the field of online convex optimization addressed the sparsity issue in the online deterministic setting, but from a quite different angle. They focus on algorithms which output sparse linear combinations, while we are interested in algorithms whose regret is small under a sparsity scenario, that is, on $\ell^0$-balls of small radii. See, for example, the papers by \citet{LaLiZh-09-TruncatedGradient}, \citet{ShaTe-11-StochasticL1Regularization}, \citet{Xia-10-DualAveragingRegularized}, \citet{DuShaSiTe-10-CompostiveObjectiveMD} and the references therein. All these articles focus on convex regularization. In the particular case of $\ell^1$-regularization under the square loss, the aforementioned works propose algorithms which predict as a sparse linear combination $\wh{y}_t = \widehat{\bu}_t \cdot \bphi(x_t)$ of the base forecasts (i.e., $\norm[\widehat{\bu}_t]_0$ is small), while no such guarantee can be proved for our algorithm SeqSEW. However they prove bounds on the $\ell^1$-regularized regret of the form
\begin{equation}
\label{eqn:chapSparse-bounds-Badshape}
\sum_{t=1}^T \Bigl( (y_t - \widehat{\bu}_t \cdot \bx_t)^2 + \lambda \norm[\widehat{\bu}_t]_1 \Bigr) \leq \inf_{\bu \in \R^d} \left\{ \sum_{t=1}^T \Bigl( (y_t - \bu \cdot \bx_t)^2 + \lambda \norm[\bu]_1 \Bigr) + \widetilde{\Delta}_{T,d}(\bu) \right\}~,
\end{equation}
for some regret term $\widetilde{\Delta}_{T,d}(\bu)$ which is suboptimal on intersections of $\ell^0$- and $\ell^1$-balls as explained below. The truncated gradient algorithm of \citet[Corollary~4.1]{LaLiZh-09-TruncatedGradient} satisfies such a regret bound\footnote{The bound stated in \citet[Corollary~4.1]{LaLiZh-09-TruncatedGradient} differs from~(\ref{eqn:chapSparse-bounds-Badshape}) in that the constant before the infimum is equal to $C=1/(1 - 2 c_d^2 \eta)$, where $c_d^2 \approx \max_{1 \leq t \leq T} \sum_{j=1}^d \phi^2_j(x_t) \leq d \norm[\bphi]^2_{\infty}$, and where a reasonable choice for $\eta$ can easily be seen to be $\eta \approx 1 / \sqrt{2 c_d^2 T}$. If the base forecasts $\phi_j(x_t)$ are dense in the sense that $c_d^2 \approx d \norm[\bphi]^2_{\infty}$, then we have $C \approx 1+\sqrt{2 c_d^2 / T}$, which yields a regret bound with leading constant $1$ as in~(\ref{eqn:chapSparse-bounds-Badshape}) and with $\widetilde{\Delta}_{T,d}(\bu)$ at least of order $\sqrt{c_d^2 T} \approx \norm[\bphi]_{\infty}\!\sqrt{d T}$.} with $\widetilde{\Delta}_{T,d}(\bu)$ at least of order $\norm[\bphi]_{\infty}\!\sqrt{d T}$ when the base forecasts $\phi_j(x_t)$ are dense in the sense that $\max_{1 \leq t \leq T} \sum_{j=1}^d \phi^2_j(x_t) \approx d \, \norm[\bphi]^2_{\infty}$. This regret bound grows as a power of and not logarithmically in $d$ as is expected for sparsity regret bounds (recall that we are interested in the case when $d \gg T$).

The three other papers mentioned above do prove (some) regret bounds with a logarithmic dependence in $d$, but these bounds do not have the dependence in $\norm[\bu]_1$ and $T$ we are looking for. For $p-1 \approx 1/(\ln d)$, the $p$-norm RDA method of \citet{Xia-10-DualAveragingRegularized} and the algorithm SMIDAS of \citet{ShaTe-11-StochasticL1Regularization}---the latter being a particular case of the algorithm COMID of \citet{DuShaSiTe-10-CompostiveObjectiveMD} specialized to the $p$-norm divergence---satisfy regret bounds of the above form~(\ref{eqn:chapSparse-bounds-Badshape}) with $\widetilde{\Delta}_{T,d}(\bu) \approx \mu \norm[\bu]_1 \sqrt{T \ln d}$, for some gradient-based constant $\mu$. Therefore, in all three cases, the function $\widetilde{\Delta}$ grows at least linearly in $\norm[\bu]_1$ and as $\sqrt{T}$. This is in contrast with the logarithmic dependence in $\norm[\bu]_1$ and the fast rate $\mathcal{O}(\ln T)$ we are looking for and prove, for example, in Corollary~\ref{cor:chapSparse-SRB-known-optimalbound}.

Note that the suboptimality of the aforementioned algorithms is specific to the goal we are pursuing, that is, prediction on $\ell^0$-balls (intersected with $\ell^1$-balls). On the contrary the rate $\norm[\bu]_1 \sqrt{T \ln d}$ is more suited and actually nearly optimal for learning on $\ell^1$-balls (see \citealt{GeYu-11alt-OnlineL1}). Moreover, the predictions output by our algorithm SeqSEW are not necessarily sparse linear combinations of the base forecasts. A question left open is thus whether it is possible to design an algorithm which both ouputs sparse linear combinations (which is statistically useful and sometimes essential for computational issues) and satisfies a sparsity regret bound of the form~(\ref{eqn:chapSparse-bounds-shape}).

\subsection{PAC-Bayesian Analysis in the Framework of Individual Sequences}

To derive our sparsity regret bounds, we follow a PAC-Bayesian approach combined with the choice of a sparsity-favoring prior. We do not have the space to review the PAC-Bayesian literature in the stochastic setting and only refer the reader to \citet{Catoni01StFlour} for a thorough introduction to the subject. As for the online deterministic setting, PAC-Bayesian-type inequalities were proved in the framework of prediction with expert advice, for example, by \citet{FrShSiWa97UsingPredictors} and \citet{KiWa99AveragingExpertPredictions}, or in the same setting as ours with a Gaussian prior by \citet{Vo01CompetitiveOnline}. More recently, \citet{Au09FastRates} proved a PAC-Bayesian result on individual sequences for general losses and prediction sets. The latter result relies on a unifying assumption called the online variance inequality, which holds true, for example, when the loss function is exp-concave. In the present paper, we only focus on the particular case of the square loss. We first use Theorem~4.6 of \citet{Au09FastRates} to derive a non-adaptive sparsity regret bound. We then provide an adaptive online PAC-Bayesian inequality to automatically adapt to the unknown range of the observations $\max_{1 \leq t \leq T} |y_t|$.

\subsection{Application to the Stochastic Setting When the Noise Level Is Unknown}
\label{sec:intro-appliSto}

In Section~\ref{sec:chapSparse-adaptivityVariance-random} we apply an automatically-tuned version of our algorithm SeqSEW on {i.i.d.}\ data. Thanks to the standard online-to-batch conversion, our sparsity regret bounds---of deterministic nature---imply a sparsity oracle inequality of the same flavor as a result of \citet{DaTsy10MirrorAveraging}. However, our risk bound holds on the whole $\R^d$ space instead of $\ell^1$-balls of finite radii, which solves one question left open by \citet[Section~4.2]{DaTsy10MirrorAveraging}. Besides, and more importantly, our algorithm does not need the a priori knowledge of the variance of the noise when the latter is Gaussian. Since the noise level is unknown in practice, adapting to it is important. This solves a second question raised by \citet[Section~5.1, Remark~6]{DaTsy10MirrorAveraging}.

\subsection{Outline of the Paper}

This paper is organized as follows.  In Section~\ref{sec:chapSparse-setting} we describe our main (deterministic) setting as well as our main notations. In Section~\ref{sec:chapSparse-sparseOnlinePrediction-deterministic} we prove the aforementioned sparsity regret bounds for our algorithm SeqSEW, first when the forecaster has access to some a priori knowledge on the observations (Sections~\ref{sec:chapSparse-onlineSEW-known} and~\ref{sec:chapSparse-onlineSEW-unknownBy}), and then when no a priori information is available (Section~\ref{sec:chapSparse-onlineSEW-unknownByBphi}), which yields a fully automatic algorithm. In Section~\ref{sec:chapSparse-adaptivityVariance} we apply the algorithm SeqSEW to two stochastic settings: the regression model with random design (Section~\ref{sec:chapSparse-adaptivityVariance-random}) and the regression model with fixed design (Section~\ref{sec:chapSparse-adaptivityVariance-fixed}). Finally the appendix contains some proofs and several useful inequalities.

\section{Setting and Notations}
\label{sec:chapSparse-setting}

\begin{figure}[h]
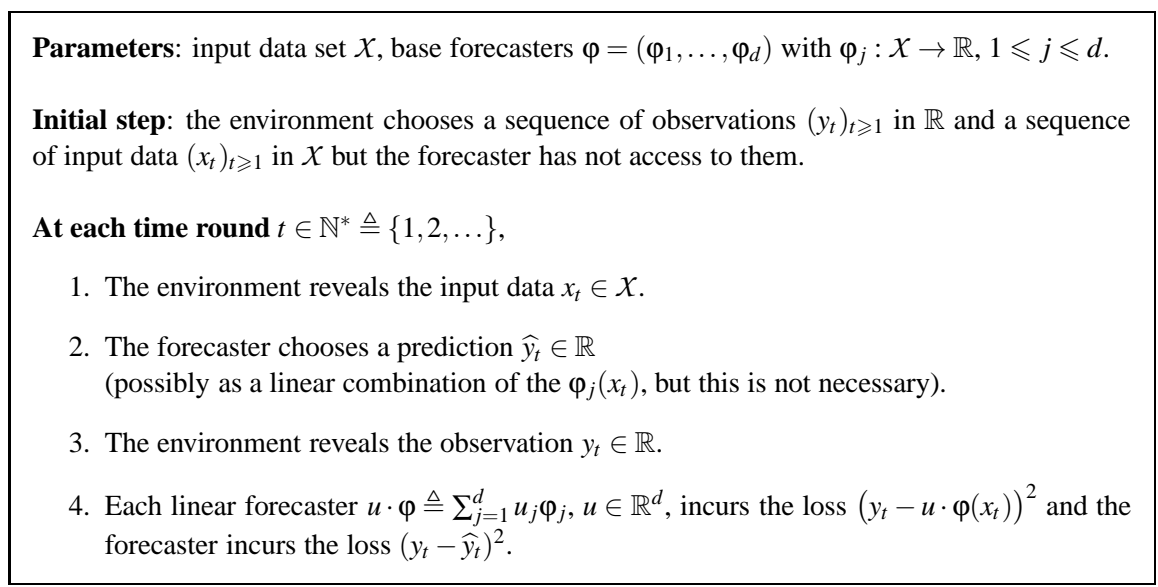

\begin{center}
\bookbox{
{\bf Parameters}: input data set $\cX$, base forecasters $\bphi = (\phi_1, \ldots, \phi_d)$ with $\phi_j: \cX \rightarrow \R$, $1 \leqslant j \leqslant d$.  \\

{\bf Initial step}: the environment chooses a sequence of observations $(y_t)_{t \geqslant 1}$ in $\R$ and a sequence of input data $(x_t)_{t \geqslant 1}$ in $\cX$ but the forecaster has not access to them. \\

{\bf At each time round} $t \in \mathbb{N}^* \eqdef \{1,2,\ldots\}$,
\begin{enumerate}
    \item The environment reveals the input data $x_t \in \cX$.
    \item The forecaster chooses a prediction $\widehat{y}_t \in \R$ \\
    (possibly as a linear combination of the $\phi_j(x_t)$, but this is not necessary).
    \item The environment reveals the observation $y_t \in \R$.
    \item Each linear forecaster $\bu \cdot \bphi \eqdef \sum_{j=1}^d u_j \phi_j$, $\bu \in \R^d$, incurs the loss $\bigl(y_t - \bu \cdot \bphi(x_t)\bigl)^2$ and the forecaster incurs the loss $(y_t - \widehat{y}_t)^2$.
\end{enumerate}
}
\end{center}
\vspace{-0.5cm}
\caption{\label{fig:chapSparse-game-sideinfo-deterministic}
The online linear regression setting.}
\end{figure}

The main setting considered in this paper is an instance of the game of prediction with expert advice called \emph{prediction with side information (under the square loss)} or, more simply, \emph{online linear regression} (see \citealt[Chapter~11]{cesa-bianchi06prediction} for an introduction to this setting). The data sequence $(x_t,y_t)_{t \geq 1}$ at hand is deterministic and arbitrary and we look for theoretical guarantees that hold for every \emph{individual} sequence. We give in Figure~\ref{fig:chapSparse-game-sideinfo-deterministic} a detailed description of our online protocol.

Note that our online protocol is described as if the environment were oblivious to the forecaster's predictions. Actually, since we only consider deterministic forecasters, all regret bounds of this paper also hold when $(x_t)_{t \geq 1}$ and $(y_t)_{t \geq 1}$ are chosen by an adversarial environment.

Two stochastic batch settings are also considered later in this paper. See Section~\ref{sec:chapSparse-adaptivityVariance-random} for the regression model with random design, and Section~\ref{sec:chapSparse-adaptivityVariance-fixed} for the regression model with fixed design.

\subsection{Some Notations}

We now define some notations. We write $\N \eqdef \{0,1,\ldots\}$ and $\rm{e} \eqdef \exp(1)$. Vectors in $\R^d$ will be denoted by bold letters. For all $\bu, \bv \in \R^d$, the standard inner product in $\R^d$ between $\bu=(u_1, \ldots, u_d)$ and $\bv=(v_1, \ldots, v_d)$ will be denoted by $\bu \cdot \bv = \sum_{i=j}^d u_j \, v_j$; the $\ell^0$-, $\ell^1$-, and $\ell^2$-norms of $\bu=(u_1, \ldots, u_d)$ are respectively defined by

\[
\norm[\bu]_0 \eqdef \sum_{j=1}^d \indic{u_j \neq 0} = \bigl|\{j: u_j \neq 0\}\bigr|~, \qquad
\norm[\bu]_1 \eqdef \sum_{j=1}^d |u_j|~, \qquad \textrm{and} \quad
\norm[\bu]_2 \eqdef \left(\sum_{j=1}^d u_j^2\right)^{1/2}~.
\]

The set of all probability distributions on a set $\Theta$ (endowed with some $\sigma$-algebra, for example, the Borel $\sigma$-algebra when $\Theta = \R^d$) will be denoted by $\cM_1^+(\Theta)$. For all $\rho, \pi \in \cM_1^+(\Theta)$, the Kullback-Leibler divergence between $\rho$ and $\pi$ is defined by

\begin{numcases}{\KL(\rho,\pi) \eqdef}
\nonumber
\int_{\R^d} \ln \left(\frac{\dd \rho}{\dd \pi}\right) \dd \rho & if $\rho$ is absolutely continuous with respect to $\pi$; \\
\nonumber
+\infty & otherwise,
\end{numcases}
where $\frac{\textrm{d} \rho}{\textrm{d} \pi}$ denotes the Radon-Nikodym derivative of $\rho$ with respect to $\pi$.

For all $x \in \R$ and $B > 0$, we denote by $\lceil x \rceil$ the smallest integer larger than or equal to $x$, and by $[x]_B$ its thresholded (or clipped) value:

\begin{numcases}{[x]_B \eqdef}
\nonumber
-B & if $x < -B$; \\
\nonumber
x & if $-B \leqslant x \leqslant B$; \\
\nonumber
B & if $x > B$.
\end{numcases}

Finally, we will use the (natural) conventions $1/0=+\infty$, $(+\infty) \times 0 = 0$, and $0 \ln (1+U/0) = 0$ for all $U \geq 0$. Any sum $\sum_{s=1}^0 a_s$ indexed from $1$ up to $0$ is by convention equal to $0$.

\section{Sparsity Regret Bounds for Individual Sequences}
\label{sec:chapSparse-sparseOnlinePrediction-deterministic}

In this section we prove sparsity regret bounds for different variants of our algorithm \mbox{SeqSEW}. We first assume in Section~\ref{sec:chapSparse-onlineSEW-known} that the forecaster has access in advance to a bound $B_y$ on the observations $|y_t|$ and a bound $B_{\Phi}$ on the trace of the empirical Gram matrix. We then remove these requirements one by one in Sections~\ref{sec:chapSparse-onlineSEW-unknownBy} and~\ref{sec:chapSparse-onlineSEW-unknownByBphi}.

\subsection{Known Bounds $B_y$ on the Observations and $B_{\Phi}$ on the Trace of the Empirical Gram Matrix}
\label{sec:chapSparse-onlineSEW-known}

To simplify the analysis, we first assume that, at the beginning of the game, the number of rounds $T$ is known to the forecaster and that he has access to a bound $B_y$ on all the observations $y_1, \ldots, y_T$ and to a bound $B_{\Phi}$ on the trace of the empirical Gram matrix, that is,

\[
y_1, \ldots, y_T \in [-B_y, B_y] \qquad \textrm{and} \qquad \sum_{j=1}^d \sum_{t=1}^T \phi_j^2(x_t) \leq B_{\Phi}~.
\]

The first version of the algorithm studied in this paper is defined in Figure~\ref{fig:chapSparse-continuouseEWA-def} (adaptive variants will be introduced later). We name it \emph{SeqSEW} for it is a variant of the Sparse Exponential Weighting algorithm introduced in the stochastic setting by \citet{DaTsy07SEW, DaTsy08SEW} which is tailored for the prediction of individual sequences.

The choice of the heavy-tailed prior $\pi_{\tau}$ is due to \citet{DaTsy07SEW}. The role of heavy-tailed priors to tackle the sparsity issue was already pointed out earlier; see, for example, the discussion by \citet[Section~2.1]{See-08-BayesianSparse}. In high dimension, such heavy-tailed priors favor sparsity: sampling from these prior distributions (or posterior distributions based on them) typically results in approximately sparse vectors, that is, vectors having most coordinates almost equal to zero and the few remaining ones with quite large values.\\[-0.3cm]

\begin{figure}[ht]
\begin{center}
\bookbox{
{\bf Parameters}: threshold $B > 0$, inverse temperature $\eta > 0$, and prior scale $\tau > 0$ with which we associate the \emph{sparsity prior} $\pi_{\tau} \in \cM_1^+(\R^d)$ defined by
\begin{equation*}
\pi_{\tau}(\d\bu) \eqdef \prod_{j=1}^d \frac{(3/\tau) \, \d u_j}{2 \bigl(1+|u_j|/\tau\bigr)^4}~.
\end{equation*}

\vspace{-0.2cm}
{\bf Initialization}: $p_1 \eqdef \pi_{\tau}$. \\

\vspace{-0.2cm}
{\bf At each time round} $t \geqslant 1$,

\vspace{-0.3cm}
\begin{enumerate}
    \item Get the input data $x_t$ and predict\footnote{The clipping operator $[\cdot]_B$ is defined in Section~\ref{sec:chapSparse-setting}.} as $\displaystyle{\wh{y}_t \eqdef \int_{\R^d} \bigl[\bu \cdot \bphi(x_t)\bigr]_B \, p_t(\d\bu)}$~;\\[-0.3cm]
    \item Get the observation $y_t$ and compute the posterior distribution $p_{t+1} \in \cM_1^+(\R^d)$ as
    \[
		p_{t+1}(\d\bu) \eqdef \frac{\exp\left(- \eta \, \displaystyle{\sum_{s=1}^t \Bigl(y_s - \bigl[\bu \cdot \bphi(x_s)\bigr]_B\Bigr)^2}\right)}{W_{t+1}} \, \pi_{\tau}(\d\bu)~,
		\]
		where
		\[
		W_{t+1} \eqdef \int_{\R^d} \exp\left(- \eta \sum_{s=1}^t \Bigl(y_s - \bigl[\bv \cdot \bphi(x_s)\bigr]_B\Bigr)^2\right) \, \pi_{\tau}(\d\bv)~.
		\]
\end{enumerate}
}
\end{center}
\vspace{-0.7cm}
\caption{\label{fig:chapSparse-continuouseEWA-def}
The algorithm $\rm{SeqSEW}^{B,\eta}_{\tau}$.}
\end{figure}

\begin{proposition}
\label{prop:chapSparse-SRB-known}
Assume that, for a known constant $B_y > 0$, the $(x_1, y_1), \ldots, (x_T, y_T)$ are such that $y_1, \ldots, y_T \in [-B_y, B_y]$. Then, for all $B \geq B_y$, all $\eta \leq 1/(8 B^2)$, and all $\tau > 0$, the algorithm $\rm{SeqSEW}^{B,\eta}_{\tau}$ satisfies
\begin{equation}
\sum_{t=1}^T (y_t - \widehat{y}_t)^2 \leq \inf_{\bu \in \R^d} \left\{\sum_{t=1}^T \bigl(y_t - \bu \cdot \bphi(x_t)\bigr)^2 + \frac{4}{\eta} \norm[\bu]_0 \ln\!\left(1 + \frac{\Arrowvert \bu \Arrowvert_1}{\norm[\bu]_0 \tau}\right)\right\} \, + \tau^2 \sum_{j=1}^d \sum_{t=1}^T \phi_j^2(x_t)~. \label{eqn:chapSparse-SRB-known-noTuning}
\end{equation}
\end{proposition}

\begin{corollary}
\label{cor:chapSparse-SRB-known-optimalbound}
Assume that, for some known constants $B_y > 0$ and $B_{\Phi} > 0$, the
$(x_1, y_1), \ldots, (x_T, y_T)$ are such that $y_1, \ldots, y_T \in [-B_y, B_y]$ and $\sum_{j=1}^d \sum_{t=1}^T \phi_j^2(x_t) \leq B_{\Phi}$~. \\
Then, when used with $B = B_y$, $\displaystyle{\eta = \frac{1}{8 B_y^2}}$, and $\displaystyle{\tau = \sqrt{\frac{16 B_y^2}{B_{\Phi}}}}$, the algorithm $\rm{SeqSEW}^{B,\eta}_{\tau}$ satisfies
\begin{equation}
\label{eqn:chapSparse-SRB-known-withTuning}
\sum_{t=1}^T (y_t - \widehat{y}_t)^2 \leq \inf_{\bu \in \R^d} \left\{\sum_{t=1}^T \bigl(y_t - \bu \cdot \bphi(x_t)\bigr)^2 + 32 \, B_y^2 \norm[\bu]_0 \ln\left(1 + \frac{\sqrt{B_{\Phi}} \, \Arrowvert \bu \Arrowvert_1}{4 \, B_y \norm[\bu]_0}\right)\right\} \, + \, 16 B_y^2~.
\end{equation}
\end{corollary}

Note that, if $\norm[\bphi]_{\infty} \eqdef \sup_{x \in \cX} \max_{1 \leq j \leq d} |\phi_j(x)|$ is finite, then the last corollary provides a \emph{sparsity regret bound} in the sense of \eqref{eqn:chapSparse-bounds-shape}. Indeed, in this case, we can take $B_{\Phi} = d \, T \norm[\bphi]_{\infty}^2$, which yields a regret bound proportional to $\norm[\bu]_0$ and that grows logarithmically in $d$, $T$, $\norm[\bu]_1$, and $\norm[\bphi]_{\infty}$.

To prove Proposition~\ref{prop:chapSparse-SRB-known}, we first need the following deterministic PAC-Bayesian inequality which is at the core of our analysis. It is a straightforward consequence of Theorem~4.6 of \citet{Au09FastRates} when applied to the square loss. An adaptive variant of this inequality will be provided in Section~\ref{sec:chapSparse-onlineSEW-unknownBy}.

\begin{lemma}
\label{lem:chapSparse-PACB-known}
Assume that for some known constant $B_y > 0$, we have $y_1, \ldots, y_T \in [-B_y, B_y]$.

For all $\tau > 0$, if the algorithm $\rm{SeqSEW}^{B,\eta}_{\tau}$ is used with $B \geq B_y$ and $\eta \leq 1/(8 B^2)$, then
\begin{align}
\sum_{t=1}^T (y_t - \widehat{y}_t)^2 \leq & \inf_{\rho \in \cM_1^+(\R^d)} \left\{\int_{\R^d} \sum_{t=1}^T \Bigl(y_t - \bigl[\bu \cdot \bphi(x_t)\bigr]_B\Bigr)^2 \rho(\dd\bu) \, + \, \frac{\KL(\rho, \pi_{\tau})}{\eta} \right\} \label{eqn:chapSparse-PACB-known-truncated} \\
\leq & \inf_{\rho \in \cM_1^+(\R^d)} \left\{\int_{\R^d} \sum_{t=1}^T \bigl(y_t - \bu \cdot \bphi(x_t)\bigr)^2 \rho(\dd\bu) \, + \, \frac{\KL(\rho, \pi_{\tau})}{\eta} \right\}~. \label{eqn:chapSparse-PACB-known-notTruncated}
\end{align}
\end{lemma}

\begin{proofref}{Lemma~\ref{lem:chapSparse-PACB-known}}
Inequality~(\ref{eqn:chapSparse-PACB-known-truncated}) is a straightforward consequence of Theorem~4.6 of \citet{Au09FastRates} when applied to the square loss, the set of prediction functions $\mathcal{G} \eqdef \bigl\{x \mapsto \bigl[\bu \cdot \bphi(x)\bigr]_B: \bu \in \R^d\bigr\}$, and the prior\footnote{The set $\mathcal{G}$ is endowed with the $\sigma$-algebra generated by all the coordinate mappings $g \in \mathcal{G} \mapsto g(x) \in \R$, $x \in \cX$ (where $\R$ is endowed with its Borel $\sigma$-algebra).} $\tilde{\pi_{\tau}}$ on $\mathcal{G}$ induced by the prior $\pi_{\tau}$ on $\R^d$ via the mapping $\bu \in \R^d \mapsto \bigl[\bu \cdot \bphi(\cdot)\bigr]_B \in \mathcal{G}$.

To apply the aforementioned theorem, recall from  \citet[Section~3.3]{cesa-bianchi06prediction} that the square loss is $1/(8 B^2)$-exp-concave on $[-B, B]$ and thus $\eta$-exp-concave,\footnote{This means that for all $y \in [-B,B]$, the function $x \mapsto \exp\bigl(- \eta (y-x)^2\bigr)$ is concave on $[-B, B]$.} since $\eta \leq 1/(8 B^2)$ by assumption. Therefore, by Theorem~4.6 of \citet{Au09FastRates} with the variance function $\delta_{\eta} \equiv 0$ (see the comments following Remark~4.1 therein), we get
\begin{align*}
\sum_{t=1}^T (y_t - \widehat{y}_t)^2 \leq & \inf_{\mu \in \cM_1^+(\mathcal{G})} \left\{\int_{\mathcal{G}} \sum_{t=1}^T \bigl(y_t - g(x_t)\bigr)^2 \mu(\dd g) \, + \, \frac{\KL(\mu, \tilde{\pi_{\tau}}\,)}{\eta} \right\} \\
\leq & \inf_{\rho \in \cM_1^+(\R^d)} \left\{\int_{\R^d} \sum_{t=1}^T \Bigl(y_t - \bigl[\bu \cdot \bphi(x_t)\bigr]_B\Bigr)^2 \rho(\dd\bu) \, + \, \frac{\KL(\tilde{\rho}, \tilde{\pi_{\tau}}\,)}{\eta} \right\}~,
\end{align*}
where the last inequality follows by restricting the infimum over $\cM_1^+\bigl(\mathcal{G}\bigr)$ to the subset $\bigl\{\tilde{\rho}: \rho \in \cM_1^+(\R^d)\bigr\} \subset \cM_1^+\bigl(\mathcal{G}\bigr)$, where $\tilde{\rho} \in \cM_1^+\bigl(\mathcal{G}\bigr)$ denotes the probability distribution induced by $\rho \in \cM_1^+(\R^d)$ via the mapping $\bu \in \R^d \mapsto \bigl[\bu \cdot \bphi(\cdot)\bigr]_B \in \mathcal{G}$. Inequality~(\ref{eqn:chapSparse-PACB-known-truncated}) then follows from the fact that for all $\rho \in \cM_1^+(\R^d)$, we have $\KL(\tilde{\rho}, \tilde{\pi_{\tau}} \,) \leq \KL(\rho, \pi_{\tau})$ by joint convexity of $\KL(\cdot,\cdot)$.

As for Inequality~(\ref{eqn:chapSparse-PACB-known-notTruncated}), it follows from (\ref{eqn:chapSparse-PACB-known-truncated}) by noting that
\[
\forall y \in [-B, B], \quad \forall x \in \R, \qquad \bigl|y - [x]_B\bigr| \leq |y - x|~.
\]
Therefore, truncation to $[-B,B]$ can only improve prediction under the square loss if the observations are $[-B,B]$-valued, which is the case here since by assumption $y_t \in [-B_y,B_y] \subset [-B,B]$ for all $t = 1, \ldots, T$.
\end{proofref}

\begin{remark}
\label{rmk:chapSparse-PACB-genericPrior}
As can be seen from the previous proof, if the prior $\pi_{\tau}$ used to define the algorithm SeqSEW was replaced with any prior $\pi \in \cM_1^+(\R^d)$, then  Lemma~\ref{lem:chapSparse-PACB-known} would still hold true with $\pi$ instead of $\pi_{\tau}$. This fact is natural from a PAC-Bayesian perspective \citep[see, e.g.,][]{Catoni01StFlour,DaTsy08SEW}. We only---but crucially---use the particular shape of the sparsity-favoring prior $\pi_{\tau}$ to derive Proposition~\ref{prop:chapSparse-SRB-known} from the PAC-Bayesian bound~\eqref{eqn:chapSparse-PACB-known-notTruncated}.
\end{remark}

\begin{proofref}{Proposition~\ref{prop:chapSparse-SRB-known}}
Our proof mimics the proof of Theorem~5 by \citet{DaTsy08SEW}. We thus only write the outline of the proof and stress the minor changes that are needed to derive Inequality~(\ref{eqn:chapSparse-SRB-known-noTuning}). The key technical tools provided by \citet{DaTsy08SEW} are reproduced in Appendix~\ref{apx:chapSparse-tools-PACB} for the convenience of the reader.

Let $\bu^* \in \R^d$. Since $B \geq B_y$ and $\eta \leq 1 / (8 B^2)$, we can apply Lemma~\ref{lem:chapSparse-PACB-known} and get
\begin{align}
\sum_{t=1}^T (y_t - \widehat{y}_t)^2 \leq & \inf_{\rho \in \cM_1^+(\R^d)} \left\{\int_{\R^d} \sum_{t=1}^T \bigl(y_t - \bu \cdot \bphi(x_t)\bigr)^2 \rho(\d\bu) \, + \, \frac{\KL(\rho, \pi_{\tau})}{\eta} \right\} \nonumber \\
\leq & \underbrace{\int_{\R^d} \sum_{t=1}^T \bigl(y_t - \bu \cdot \bphi(x_t)\bigr)^2 \rho_{\bu^*,\tau}(\d\bu)}_{(1)} \, + \, \underbrace{\frac{\KL(\rho_{\bu^*,\tau}, \pi_{\tau})}{\eta}}_{(2)}~. \label{eqn:chapSparse-SRB-known-PACBupperbound}
\end{align}
In the last inequality, $\rho_{\bu^*,\tau}$ is taken as the translated of $\pi_{\tau}$ at $\bu^*$, namely,
\[
\rho_{\bu^*,\tau}(\d\bu) \eqdef \frac{\d\pi_{\tau}}{\d\bu} (\bu - \bu^*) \, \d\bu = \prod_{j=1}^d \frac{(3/\tau) \, \d u_j}{2 \bigl(1+|u_j - u^*_j|/\tau\bigr)^4}~.
\]
The two terms $(1)$ and $(2)$ can be upper bounded as in the proof of Theorem~5 by \citet{DaTsy08SEW}. By a symmetry argument recalled in Lemma~\ref{lem:chapSparse-PACBupperbound-int} (Appendix~\ref{apx:chapSparse-tools-PACB}), the first term $(1)$ can be rewritten as
\begin{equation}
\label{eqn:chapSparse-SRB-known-PACBupperbound1}
\int_{\R^d} \sum_{t=1}^T \bigl(y_t - \bu \cdot \bphi(x_t)\bigr)^2 \rho_{\bu^*,\tau}(\d\bu) = \sum_{t=1}^T \bigl(y_t - \bu^* \cdot \bphi(x_t)\bigr)^2 + \tau^2 \sum_{j=1}^d \sum_{t=1}^T \phi_j^2(x_t)~.
\end{equation}
As for the term $(2)$, we have, as is recalled in Lemma~\ref{lem:chapSparse-PACBupperbound-KL},
\begin{equation}
\frac{\KL(\rho_{\bu^*,\tau}, \pi_{\tau})}{\eta} \leq \frac{4}{\eta} \norm[\bu^*]_0 \ln\left(1 + \frac{\norm[\bu^*]_1}{\norm[\bu^*]_0 \tau}\right)~. \label{eqn:chapSparse-SRB-known-PACBupperbound2}
\end{equation}
Combining (\ref{eqn:chapSparse-SRB-known-PACBupperbound}), (\ref{eqn:chapSparse-SRB-known-PACBupperbound1}), and (\ref{eqn:chapSparse-SRB-known-PACBupperbound2}), which all hold for all $\bu^* \in \R^d$, we get Inequality~(\ref{eqn:chapSparse-SRB-known-noTuning}).
\end{proofref}

\begin{proofref}{Corollary~\ref{cor:chapSparse-SRB-known-optimalbound}}
Applying Proposition~\ref{prop:chapSparse-SRB-known}, we have, since $B \geq B_y$ and $\eta \leq 1 / (8 B^2)$,
\begin{align}
\sum_{t=1}^T (y_t - \widehat{y}_t)^2 \leq & \inf_{\bu \in \R^d} \!\left\{\sum_{t=1}^T \bigl(y_t - \bu \cdot \bphi(x_t)\bigr)^2 + \frac{4}{\eta} \norm[\bu]_0 \ln\!\left(1 \!+ \!\frac{\norm[\bu]_1}{\norm[\bu]_0 \tau}\right)\!\right\} + \tau^2 \sum_{j=1}^d \sum_{t=1}^T \phi_j^2(x_t) \nonumber \\
\leq & \inf_{\bu \in \R^d} \!\left\{\sum_{t=1}^T \bigl(y_t - \bu \cdot \bphi(x_t)\bigr)^2 + \frac{4}{\eta} \norm[\bu]_0 \ln\!\left(1 + \frac{\norm[\bu]_1}{\norm[\bu]_0 \tau}\right)\right\} + \tau^2 B_{\Phi}~, \nonumber 
\end{align}
since $\sum_{j=1}^d \sum_{t=1}^T \phi_j^2(x_t) \leq B_{\Phi}$ by assumption. The particular (and nearly optimal) choices of $\eta$ and $\tau$ given in the statement of the corollary then yield the desired inequality~(\ref{eqn:chapSparse-SRB-known-withTuning}).
\end{proofref}

We end this subsection with a natural question about approximate sparsity: Proposition~\ref{prop:chapSparse-SRB-known} ensures a low regret with respect to sparse linear combinations $\bu \cdot \bphi$, but what can be said for approximately sparse linear combinations, that is, predictors of the form $\bu \cdot \bphi$ where $\bu \in \R^d$ is very close to a sparse vector? As can be seen from the proof of Lemma~\ref{lem:chapSparse-PACBupperbound-KL} in Appendix~\ref{apx:chapSparse-tools-PACB}, the sparsity-related term
\[
\frac{4}{\eta} \norm[\bu]_0 \ln\!\left(1 + \frac{\Arrowvert \bu \Arrowvert_1}{\norm[\bu]_0 \tau}\right)
\]
in the regret bound of Proposition~\ref{prop:chapSparse-SRB-known} can actually be replaced with the smaller (and continous) term
\[
\frac{4}{\eta} \, \sum_{j=1}^d \ln\left(1+|u_j|/\tau\right)~.
\]
The last term is always smaller than the former and guarantees that the regret is small with respect to any approximately sparse vector $\bu \in \R^d$.

\subsection{Unknown Bound $B_y$ on the Observations but Known Bound $B_{\Phi}$ on the Trace of the Empirical Gram Matrix}
\label{sec:chapSparse-onlineSEW-unknownBy}

In the previous section, to prove the upper bounds stated in Lemma~\ref{lem:chapSparse-PACB-known} and Proposition~\ref{prop:chapSparse-SRB-known}, we assumed that the forecaster had access to a bound $B_y$ on the observations $|y_t|$ and to a bound $B_{\Phi}$ on the trace of the empirical Gram matrix. In this section, we remove the first requirement and prove a sparsity regret bound for a variant of the algorithm $\rm{SeqSEW}^{B,\eta}_{\tau}$ which is adaptive to the unknown bound $B_y=\max_{1 \leq t \leq T} |y_t|$; see Proposition~\ref{prop:chapSparse-SRB-unknownBy} and Remark~\ref{rmk:COLT11-SRB-unknownBy-adaptivity} below.

\begin{figure}[ht]
\hspace{-0.2cm}
\begin{center}
\bookbox{
{\bf Parameter}: prior scale $\tau > 0$ with which we associate the \emph{sparsity prior} $\pi_{\tau} \in \cM_1^+(\R^d)$ defined by
\begin{equation*}
\pi_{\tau}(\d\bu) \eqdef \prod_{j=1}^d \frac{(3/\tau) \, \d u_j}{2 \bigl(1+|u_j|/\tau\bigr)^4}~.
\end{equation*}

{\bf Initialization}: $B_1 \eqdef 0$, $\eta_1 \eqdef +\infty$, and $p_1 \eqdef \pi_{\tau}$. \\

\vspace{-0.2cm}
{\bf At each time round} $t \geqslant 1$,
\begin{enumerate}
    \vspace{-0.2cm}
    \item Get the input data $x_t$ and predict\footnote{The clipping operator $[\cdot]_B$ is defined in Section~\ref{sec:chapSparse-setting}.} as $\displaystyle{\wh{y}_t \eqdef \int_{\R^d} \bigl[\bu \cdot \bphi(x_t)\bigr]_{B_t} \, p_t(\d\bu)}$;
    \vspace{-0.1cm}
    \item Get the observation $y_t$ and update:
    \begin{itemize}
	\vspace{-0.2cm}
	\item the threshold $B_{t+1} \eqdef \max_{1 \leq s \leq t} |y_s|$,
	\vspace{-0.1cm}
	\item the inverse temperature $\eta_{t+1} \eqdef 1/\bigl(8 B_{t+1}^2\bigr)$~,
	\item and the posterior distribution $p_{t+1} \in \cM_1^+(\R^d)$ as
    \[
		p_{t+1}(\d\bu) \eqdef \frac{\exp\left(- \eta_{t+1} \, \displaystyle{\sum_{s=1}^t \Bigl(y_s - \bigl[\bu \cdot \bphi(x_s)\bigr]_{B_s}\Bigr)^2}\right)}{W_{t+1}} \, \pi_{\tau}(\d\bu)~,
		\]
		where
\vspace{-0.5cm}
		\[
		W_{t+1} \eqdef \int_{\R^d} \exp\left(- \eta_{t+1} \sum_{s=1}^t \Bigl(y_s - \bigl[\bv \cdot \bphi(x_s)\bigr]_{B_s}\Bigr)^2\right) \, \pi_{\tau}(\d\bv)~.
		\]
		\end{itemize}
\end{enumerate}
}
\end{center}
\vspace{-0.5cm}
\caption{\label{fig:chapSparse-continuouseEWA-varyingeta-def}
The algorithm $\rm{SeqSEW}^*_{\tau}$.}
\end{figure}

For this purpose we consider the algorithm of Figure~\ref{fig:chapSparse-continuouseEWA-varyingeta-def}, which we call $\rm{SeqSEW}^*_{\tau}$ thereafter. It differs from $\rm{SeqSEW}^{B,\eta}_{\tau}$ defined in the previous section in that the threshold $B$ and the inverse temperature $\eta$ are now allowed to vary over time and are chosen at each time round as a function of the data available to the forecaster.

The idea of truncating the base forecasts was used many times in the past; see, for example, the work of \citet{Vo01CompetitiveOnline} in the online linear regression setting, that of \citet[Chapter~10]{GyKoKrWa-02-DistributionFreeNonparametric} for the regression problem with random design, and the papers of \citet{GyOt07SeqPredictionUnbounded} and \citet{BiBlGyOt-10-NonparametricSeqPrediction} for sequential prediction of unbounded time series under the square loss. A key ingredient in the present paper is to perform truncation with respect to a data-driven threshold.

\begin{proposition}
\label{prop:chapSparse-SRB-unknownBy}
For all $\tau > 0$, the algorithm $\rm{SeqSEW}^*_{\tau}$ satisfies
\begin{align}
\sum_{t=1}^T (y_t - \widehat{y}_t)^2 \leq & \inf_{\bu \in \R^d} \left\{\sum_{t=1}^T \bigl(y_t - \bu \cdot \bphi(x_t)\bigr)^2 + 32 B_{T+1}^2 \norm[\bu]_0 \ln\left(1 + \frac{\norm[\bu]_1}{\norm[\bu]_0 \tau}\right)\right\} \nonumber \\ 
& + \, \tau^2 \sum_{j=1}^d \sum_{t=1}^T \phi_j^2(x_t)  \, + 5 B_{T+1}^2~, \nonumber
\end{align}
where $B_{T+1}^2 \eqdef \max_{1 \leq t \leq T} y_t^2$.
\end{proposition}

\begin{remark}
\label{rmk:COLT11-SRB-unknownBy-adaptivity}
{\em
In view of Proposition~\ref{prop:chapSparse-SRB-known}, the algorithm $\rm{SeqSEW}^*_{\tau}$ satisfies a sparsity regret bound which is adaptive to the unknown bound $B_y = \max_{1 \leq t \leq T} |y_t|$. The price for the automatic tuning with respect to $B_y$ consists only of the additive term $5 B_{T+1}^2 = 5 B_y^2$.
}
\end{remark}

As in the previous section, several corollaries can be derived from Proposition~\ref{prop:chapSparse-SRB-unknownBy}. If the forecaster has access beforehand to a quantity $B_{\Phi} > 0$ such that $\sum_{j=1}^d \sum_{t=1}^T \phi_j^2(x_t) \leq B_{\Phi}$, then a suboptimal but reasonable choice of $\tau$ is given by $\tau = 1/ \sqrt{B_{\Phi}}$; see Corollary~\ref{cor:chapSparse-SRB-unknownBy-tauOptimal} below. The simpler tuning $\tau = 1/ \sqrt{d T}$ of Corollary~\ref{cor:chapSparse-SRB-unknownBy-tauSimple} will be useful in the stochastic batch setting ({cf.}, Section~\ref{sec:chapSparse-adaptivityVariance}).\footnote{The tuning $\tau = 1/ \sqrt{d T}$ only uses the knowledge of $T$, which is known by the forecaster in the stochastic batch setting. In that framework, another simple and easy-to-analyse tuning is given by $\tau = 1/(\norm[\bphi]_{\infty} \!\sqrt{d \, T})$---which corresponds to $B_{\Phi} = d \, T \norm[\bphi]_{\infty}^2$---but it requires that $\norm[\bphi]_{\infty} \eqdef \sup_{x \in \cX} \max_{1 \leq j \leq d} |\phi_j(x)|$ be finite. Note that the last tuning satisfies the scale-invariant property pointed out by \citet[Remark~4]{DaTsy10MirrorAveraging}.} The proofs of the next corollaries are immediate.

\begin{corollary}
\label{cor:chapSparse-SRB-unknownBy-tauOptimal}
Assume that, for a known constant $B_{\Phi} > 0$, the $(x_1, y_1), \ldots, (x_T, y_T)$ are such that $\sum_{j=1}^d \sum_{t=1}^T \phi_j^2(x_t) \leq B_{\Phi}$. Then, when used with $\tau = 1/\sqrt{B_{\Phi}}$, the algorithm $\rm{SeqSEW}^*_{\tau}$ satisfies
\begin{align}
\sum_{t=1}^T (y_t - \widehat{y}_t)^2 \leq & \inf_{\bu \in \R^d} \left\{\sum_{t=1}^T \bigl(y_t - \bu \cdot \bphi(x_t)\bigr)^2 + 32 B_{T+1}^2 \norm[\bu]_0 \ln\left(1 + \frac{\sqrt{B_{\Phi}} \, \norm[\bu]_1}{\norm[\bu]_0}\right)\right\} \nonumber \\ 
& + 5 B_{T+1}^2 + 1~, \nonumber
\end{align}
where $B_{T+1}^2 \eqdef \max_{1 \leq t \leq T} y_t^2$.
\end{corollary}

\begin{corollary}
\label{cor:chapSparse-SRB-unknownBy-tauSimple}
Assume that $T$ is known to the forecaster at the beginning of the prediction game. Then, when used with $\tau=1/\sqrt{d T}$,
the algorithm $\rm{SeqSEW}^*_{\tau}$ satisfies
\begin{align}
\sum_{t=1}^T (y_t - \widehat{y}_t)^2 \leq & \inf_{\bu \in \R^d} \left\{\sum_{t=1}^T \bigl(y_t - \bu \cdot \bphi(x_t)\bigr)^2 + 32 B_{T+1}^2 \norm[\bu]_0 \ln\left(1 + \frac{\sqrt{d T} \, \norm[\bu]_1}{\norm[\bu]_0}\right)\right\} \nonumber \\ 
& + \frac{1}{d T} \sum_{j=1}^d \sum_{t=1}^T \phi_j^2(x_t) + 5 B_{T+1}^2~, \nonumber
\end{align}
where $B_{T+1}^2 \eqdef \max_{1 \leq t \leq T} y_t^2$.
\end{corollary}

As in the previous section, to prove Proposition~\ref{prop:chapSparse-SRB-unknownBy}, we first need a key PAC-Bayesian inequality. The next lemma is an adaptive variant of Lemma~\ref{lem:chapSparse-PACB-known}.

\begin{lemma}
\label{lem:chapSparse-PACB-unknownBy}
For all $\tau > 0$, the algorithm $\rm{SeqSEW}^*_{\tau}$ satisfies
\begin{align}
\sum_{t=1}^T (y_t - \widehat{y}_t)^2 \leq & \inf_{\rho \in \cM_1^+(\R^d)} \left\{\int_{\R^d} \sum_{t=1}^T \Bigl(y_t - \bigl[\bu \cdot \bphi(x_t)\bigr]_{B_t}\Bigr)^2 \rho(\dd\bu) \, + 8 B_{T+1}^2 \, \KL(\rho, \pi_{\tau}) \right\} + 4 B_{T+1}^2 \label{eqn:chapSparse-PACB-unknownBy-truncated} \\[-0.2cm]
\leq & \inf_{\rho \in \cM_1^+(\R^d)} \left\{\int_{\R^d} \sum_{t=1}^T \bigl(y_t - \bu \cdot \bphi(x_t)\bigr)^2 \rho(\dd\bu) \, + 8 B_{T+1}^2 \, \KL(\rho, \pi_{\tau})\right\}  \, + 5 B_{T+1}^2~, \label{eqn:chapSparse-PACB-unknownBy-notTruncated}
\end{align}
\ \\[-0.5cm]
where $B_{T+1}^2 \eqdef \max_{1 \leq t \leq T} y_t^2$.
\end{lemma}

\begin{proofref}{Lemma~\ref{lem:chapSparse-PACB-unknownBy}}
The proof is based on arguments that are similar to those underlying Lemma~\ref{lem:chapSparse-PACB-known}, except that we now need to deal with $B$ and $\eta$ changing over time. In the same spirit as in \citet{AuCeGe02Adaptive}, \citet{CeMaSt07SecOrder} and \citet{GyOt07SeqPredictionUnbounded}, our analysis relies on the control of $(\ln W_{t+1})/\eta_{t+1} - (\ln W_t)/\eta_t$ where $W_1 \eqdef 1$ and, for all $t \geq 2$,
\[
W_t \eqdef \int_{\R^d} \exp\left(- \eta_t \sum_{s=1}^{t-1} \Bigl(y_s - \bigl[\bu \cdot \bphi(x_s)\bigr]_{B_s}\Bigr)^2\right) \, \pi_{\tau}(\d\bu)~.
\]

Before controlling $(\ln W_{t+1})/\eta_{t+1} - (\ln W_t)/\eta_t$, we first need a little comment. Note that all $\eta_t$'s such that $\eta_t = + \infty$ (i.e., $B_t = 0$) can be replaced with any finite value without changing the predictions of the algorithm (since the sum $\sum_{s=1}^{t-1}$ above equals zero). Therefore, we assume in the sequel that $(\eta_t)_{t \geq 1}$ is a non-decreasing sequence of \emph{finite} positive real numbers.\\

\noindent
{\em First step}: On the one hand, we have
\begin{align}
& \frac{\ln W_{T+1}}{\eta_{T+1}} - \frac{\ln W_1}{\eta_1} = \frac{1}{\eta_{T+1}} \ln \int_{\R^d} \exp\left(- \eta_{T+1} \sum_{t=1}^{T} \Bigl(y_t - \bigl[\bu \cdot \bphi(x_t)\bigr]_{B_t}\Bigr)^2\right) \pi_{\tau}(\d\bu) \, - \frac{1}{\eta_1} \ln 1 \nonumber \\
& \qquad = - \inf_{\rho \in \cM_1^+(\R^d)} \left\{\int_{\R^d} \sum_{t=1}^T \Bigl(y_t - \bigl[\bu \cdot \bphi(x_t)\bigr]_{B_t}\Bigr)^2 \rho(\d\bu)\, + \, \frac{\KL(\rho, \pi_{\tau})}{\eta_{T+1}} \right\}~, \label{eqn:chapSparse-PACB-unknownBy-truncated-minus}
\end{align}
where the last equality follows from a convex duality argument for the Kullback-Leibler divergence (cf., e.g., \citealt[p. 159]{Catoni01StFlour}) which we recall in Proposition~\ref{prop:appendix-dualityKL} in Appendix~\ref{apx:dualityKL}. \\

\noindent
{\em Second step}: On the other hand, we can rewrite $(\ln W_{T+1})/\eta_{T+1} - (\ln W_1)/\eta_1$ as a telescopic sum and get
\begin{equation}
\label{eqn:chapSparse-PACB-unknownBy-spliteta}
\frac{\ln W_{T+1}}{\eta_{T+1}} - \frac{\ln W_1}{\eta_1} = \sum_{t=1}^T \left(\frac{\ln W_{t+1}}{\eta_{t+1}} - \frac{\ln W_t}{\eta_t}\right) =  \sum_{t=1}^T \biggl(\underbrace{\frac{\ln W_{t+1}}{\eta_{t+1}} - \frac{\ln W'_{t+1}}{\eta_t}}_{(1)} + \underbrace{\frac{1}{\eta_t} \ln \frac{W'_{t+1}}{W_t}}_{(2)}\biggr)~,
\end{equation}
where $W'_{t+1}$ is obtained from $W_{t+1}$ by replacing $\eta_{t+1}$ with $\eta_t$; namely,
\[
W'_{t+1} \eqdef \int_{\R^d} \exp\left(- \eta_t \sum_{s=1}^{t} \Bigl(y_s - \bigl[\bu \cdot \bphi(x_s)\bigr]_{B_s}\Bigr)^2\right) \, \pi_{\tau}(\d\bu)~.
\]
Let $t \in \{1, \ldots, T\}$. The first term $(1)$ is non-positive by Jensen's inequality (note that $x \mapsto x^{\eta_{t+1}/\eta_t}$ is concave on $\R^*_+$ since $\eta_{t+1} \leq \eta_t$ by construction). As for the second term $(2)$, by definition of $W_{t+1}'$,
\begin{align}
& \frac{1}{\eta_t} \ln \frac{W'_{t+1}}{W_t} \nonumber \\[-0.3cm]
& \quad = \frac{1}{\eta_t} \ln \!\bigintss_{\R^d} \frac{\exp\!\left(\!- \eta_t \Bigl(y_t - \bigl[\bu \cdot \bphi(x_t)\bigr]_{B_t}\Bigr)^2\right) \exp\!\left(\!- \eta_t \displaystyle{\sum_{s=1}^{t-1} \Bigl(y_s - \bigl[\bu \cdot \bphi(x_s)\bigr]_{B_s}\Bigr)^2}\right)}{W_t} \, \pi_{\tau}(\d\bu) \nonumber \\
& \quad = \frac{1}{\eta_t} \ln \int_{\R^d} \exp\left(- \eta_t \Bigl(y_t - \bigl[\bu \cdot \bphi(x_t)\bigr]_{B_t}\Bigr)^2\right) \, p_t(\d\bu)~. \label{eqn:chapSparse-PACB-unknownBy-logLaplace1}
\end{align}
where (\ref{eqn:chapSparse-PACB-unknownBy-logLaplace1}) follows by definition of $p_t$. The next paragraphs are dedicated to upper bounding the last integral above. First note that this is straightforward in the particular case where $y_t \in [-B_t, B_t]$. Indeed, by definition of $\eta_t \eqdef 1/(8 B_t^2)$ and by the fact that the square loss is $1/(8 B_t^2)$-exp-concave on $[-B_t,B_t]$ (as in Lemma~\ref{lem:chapSparse-PACB-known}),\footnote{\label{ftn:details_eta_t}To be more exact, we assigned some arbitrary finite value to $\eta_t$ when $B_t=0$. However, in this case, the square loss is of course $\eta_t$-exp-concave on $[-B_t,B_t]=\{0\}$ whatever the value of $\eta_t$.} we get from Jensen's inequality that
\[
\int_{\R^d} e^{- \eta_t \bigl(y_t - \bigl[\bu \cdot \bphi(x_t)\bigr]_{B_t}\bigr)^2} p_t(\d\bu) \leq \exp\!\left(\!- \eta_t \left(y_t - \int_{\R^d} \bigl[\bu \cdot \bphi(x_t)\bigr]_{B_t} p_t(\d\bu)\right)^2\right) = e^{- \eta_t \left(y_t - \widehat{y}_t\right)^2},
\]
where the last equality follows by definition of $\widehat{y}_t$. Taking the logarithms of both sides of the last inequality and dividing by $\eta_t$, we can see that the quantity on the right-hand side of~\eqref{eqn:chapSparse-PACB-unknownBy-logLaplace1} is bounded from above by $-\bigl(y_t-\hat{y}_t\bigr)^2$.

In the general case, we cannot assume that $y_t \in [-B_t, B_t]$, since it may happen that $|y_t| > \max_{1 \leq s \leq t-1} |y_s| \eqdef B_t$. As shown below, we can still use the exp-concavity of the square loss if we replace $y_t$ with its clipped version $[y_t]_{B_t}$. More precisely, setting $\hat{y}_{t,\bu} \eqdef [\bu \cdot \bphi(x_t)]_{B_t}$ for all $\bu \in \R^d$, the square loss appearing in the right-hand side of \eqref{eqn:chapSparse-PACB-unknownBy-logLaplace1} equals
\begin{align}
\bigl(y_t - \hat{y}_{t,\bu} \bigr)^2 & = \bigl([y_t]_{B_t} - \hat{y}_{t,\bu}\bigr)^2 + \bigl(y_t - [y_t]_{B_t}\bigr)^2 + 2 \bigl(y_t - [y_t]_{B_t}\bigr) \bigl([y_t]_{B_t} - \hat{y}_{t,\bu}\bigr) \nonumber \\
& = \bigl([y_t]_{B_t} - \hat{y}_{t,\bu}\bigr)^2 + \bigl(y_t - [y_t]_{B_t}\bigr)^2 + 2 \bigl(y_t - [y_t]_{B_t}\bigr) \bigl([y_t]_{B_t} - \hat{y}_{t}\bigr) + c_{t,\bu}~, \label{eqn:chapSparse-PACB-unknownBy-newproof-decomposition}
\end{align}
where we set
\begin{align}
c_{t,\bu} & \eqdef 2 \bigl(y_t - [y_t]_{B_t}\bigr) \bigl(\hat{y}_{t}-\hat{y}_{t,\bu}\bigr) \nonumber \\
& \geq - 4 B_t \bigl| y_t - [y_t]_{B_t} \bigr| \geq - 4 B_t (B_{t+1}-B_t)~, \label{eqn:chapSparse-PACB-unknownBy-newproof-lowerbound}
\end{align}
where the last two inequalities follow from the property $\hat{y}_t , \hat{y}_{t,\bu} \in [-B_t,B_t]$ (by construction) and from the elementary\footnote{To see why this is true, it suffices to rewrite $[y_t]_{B_t}$ in the three cases $y_t < -B_t$, $|y_t| \leq B_t$, or $y_t > B_t$.} yet useful upper bound $\bigl| y_t - [y_t]_{B_t} \bigr| \leq B_{t+1} - B_t$.

Combining \eqref{eqn:chapSparse-PACB-unknownBy-newproof-decomposition} with the lower bound~\eqref{eqn:chapSparse-PACB-unknownBy-newproof-lowerbound} yields that, for all $\bu \in \R^d$,
\begin{align}
\bigl(y_t - \hat{y}_{t,\bu} \bigr)^2 & \geq \bigl([y_t]_{B_t} - \hat{y}_{t,\bu}\bigr)^2 + C_t~, \label{eqn:chapSparse-PACB-unknownBy-newproof-lowerbound-continued}
\end{align}
where we set $C_t \eqdef \bigl(y_t - [y_t]_{B_t}\bigr)^2 + 2 \bigl(y_t - [y_t]_{B_t}\bigr) \bigl([y_t]_{B_t} - \hat{y}_{t}\bigr) - 4 B_t (B_{t+1}-B_t)$.

We can now continue the upper bounding of $(1/\eta_t) \ln (W'_{t+1} / W_t)$. Indeed, substituting the lower bound \eqref{eqn:chapSparse-PACB-unknownBy-newproof-lowerbound-continued} into \eqref{eqn:chapSparse-PACB-unknownBy-logLaplace1}, we get that
\begin{align}
\frac{1}{\eta_t} \ln \frac{W'_{t+1}}{W_t} & \leq \frac{1}{\eta_t} \ln \!\left[\, \int_{\R^d} \exp\left(- \eta_t \bigl([y_t]_{B_t} - \hat{y}_{t,\bu} \bigr)^2\right) \, p_t(\d\bu)\right] - C_t \nonumber \\
& \leq \frac{1}{\eta_t} \ln \!\left[\exp\!\left(\!- \eta_t \left([y_t]_{B_t} - \int_{\R^d} \hat{y}_{t,\bu} \, p_t(\d\bu)\right)^2\right)\right] - C_t \label{eqn:chapSparse-PACB-unknownBy-logLaplace-continued1} \\
& = -\bigl([y_t]_{B_t}-\hat{y}_t\bigr)^2 - C_t \label{eqn:chapSparse-PACB-unknownBy-logLaplace-continued2} \\
& = -\bigl(y_t-\hat{y}_t\bigr)^2 + 4 B_t (B_{t+1}-B_t)~, \label{eqn:chapSparse-PACB-unknownBy-logLaplace-continued3}
\end{align}
where \eqref{eqn:chapSparse-PACB-unknownBy-logLaplace-continued1} follows by Jensen's inequality (recall that $\eta_t \eqdef 1/(8 B_t^2)$ and that the square loss is $1/(8 B_t^2)$-exp-concave on $[-B_t,B_t]$),\footnote{Same remark as in Footnote~\ref{ftn:details_eta_t}.} where \eqref{eqn:chapSparse-PACB-unknownBy-logLaplace-continued2} is entailed by definition of $\hat{y}_{t,\bu}$ and $\hat{y}_t$, and where \eqref{eqn:chapSparse-PACB-unknownBy-logLaplace-continued3} follows by definition of $C_t$ above and by elementary calculations.

Summing \eqref{eqn:chapSparse-PACB-unknownBy-logLaplace-continued3} over $t=1, \ldots, T$ and using the upper bound $B_t (B_{t+1}-B_t) \leq B_{t+1}^2 - B_t^2$, Equation~(\ref{eqn:chapSparse-PACB-unknownBy-spliteta}) yields
\begin{align}
\frac{\ln W_{T+1}}{\eta_{T+1}} - \frac{\ln W_1}{\eta_1} & \leq - \sum_{t=1}^T (y_t - \widehat{y}_t)^2 + 4 \sum_{t=1}^T \bigl(B_{t+1}^2 -B_t^2\bigr) \nonumber \\
& = - \sum_{t=1}^T (y_t - \widehat{y}_t)^2 + 4 B_{T+1}^2~. \label{eqn:chapSparse-PACB-unknownBy-upperbound1}
\end{align}

\noindent
{\em Third step}: Putting \eqref{eqn:chapSparse-PACB-unknownBy-truncated-minus} and \eqref{eqn:chapSparse-PACB-unknownBy-upperbound1} together, we get the PAC-Bayesian inequality
\begin{equation*}
\sum_{t=1}^T (y_t - \widehat{y}_t)^2 \leq \inf_{\rho \in \cM_1^+(\R^d)} \left\{\int_{\R^d} \sum_{t=1}^T \Bigl(y_t - \bigl[\bu \cdot \bphi(x_t)\bigr]_{B_t}\Bigr)^2 \rho(\d\bu)\, + \, \frac{\KL(\rho, \pi_{\tau})}{\eta_{T+1}} \right\} \, + 4 B_{T+1}^2~,
\end{equation*}
which yields \eqref{eqn:chapSparse-PACB-unknownBy-truncated} since $\eta_{T+1} \eqdef 1/(8 B_{T+1}^2)$ by definition.\footnote{If $B_{T+1}=0$, then $y_t=\hat{y}_t=0$ for all $1 \leq t \leq T$, which immediately yields \eqref{eqn:chapSparse-PACB-unknownBy-truncated}.} The other PAC-Bayesian inequality~\eqref{eqn:chapSparse-PACB-unknownBy-notTruncated}, which is stated for non-truncated base forecasts, is a direct consequence of \eqref{eqn:chapSparse-PACB-unknownBy-truncated} and of the following two arguments: for all $\bu \in \R^d$ and all $t=1,\ldots,T$,
\begin{align}
\bigl(y_t - [\bu \cdot \bphi(x_t)]_{B_t}\bigr)^2 & \leq \bigl(y_t - \bu \cdot \bphi(x_t) \bigr)^2 + (B_{t+1}-B_t)^2 \label{eqn:chapSparse-PACB-unknownBy-upperbound2-a}
\end{align}
and
\begin{align}
\sum_{t=1}^T (B_{t+1}-B_t)^2 \leq B_{T+1}^2~. \label{eqn:chapSparse-PACB-unknownBy-upperbound2-b}
\end{align}

\noindent
{\em Complement}: proof of \eqref{eqn:chapSparse-PACB-unknownBy-upperbound2-a} and \eqref{eqn:chapSparse-PACB-unknownBy-upperbound2-b}.\\
To see why \eqref{eqn:chapSparse-PACB-unknownBy-upperbound2-a} is true, we can distinguish between several cases. First note that this inequality is straightforward when $|y_t| \leq B_t$ (indeed, in this case, clipping $\bu \cdot \bphi(x_t)$ to $[-B_t,B_t]$ can only improve prediction). We can thus assume that $|y_t| > B_t$, or just\footnote{If $y_t < -B_t$, it suffices to apply \eqref{eqn:chapSparse-PACB-unknownBy-upperbound2-a} with $-y_t$ and $-\bu$.} that $y_t > B_t$. In this case, we can distinguish between three sub-cases:
\begin{itemize}
	\item if $\bu \cdot \bphi(x_t) < -B_t$, then clipping improves prediction since $y_t > B_t$;
	\item if $-B_t \leq \bu \cdot \bphi(x_t) \leq B_t$, then the clipping operator $[\cdot]_B$ has no effect on $\bu \cdot \bphi(x_t)$;
	\item if $\bu \cdot \bphi(x_t) > B_t$, then $[\bu \cdot \bphi(x_t)]_{B_t} = B_t$ so that $(y_t-[\bu \cdot \bphi(x_t)]_{B_t})^2 = (B_{t+1}-B_t)^2$ since $B_{t+1} = y_t$.
\end{itemize}
Therefore, in all three sub-cases described above, we have
\[
(y_t-[\bu \cdot \bphi(x_t)]_{B_t})^2 \leq \max \left\{ (y_t-\bu \cdot \bphi(x_t))^2, \, (B_{t+1}-B_t)^2 \right\}~,
\]
which concludes the proof of~\eqref{eqn:chapSparse-PACB-unknownBy-upperbound2-a}. As for~\eqref{eqn:chapSparse-PACB-unknownBy-upperbound2-b}, it follows from the inequality
\[
\sum_{t=1}^T (B_{t+1}-B_t)^2 \leq \sup_{\substack{\Delta_1,\ldots,\Delta_T \geq 0 \\ \sum_{t=1}^T \Delta_t = B_{T+1}}} \!\left\{ \sum_{t=1}^T \Delta_t^2 \right\} = B_{T+1}^2~,
\]
where the last equality is entailed by convexity of the function $(\Delta_1,\ldots,\Delta_T) \mapsto \sum_{t=1}^T \Delta_t^2$ on the polytope $\bigl\{ (\Delta_1,\ldots,\Delta_T) \in \R_+^T : \; \sum_{t=1}^T \Delta_t = B_{T+1}\bigr\}$. This concludes the proof.
\end{proofref}

\begin{proofref}{Proposition~\ref{prop:chapSparse-SRB-unknownBy}}
The proof follows exactly the same lines as in Proposition~\ref{prop:chapSparse-SRB-known}  except that we apply Lemma~\ref{lem:chapSparse-PACB-unknownBy} instead of Lemma~\ref{lem:chapSparse-PACB-known}.
Indeed, using Lemma~\ref{lem:chapSparse-PACB-unknownBy} and restricting the infimum to the $\rho_{\bu^*,\tau}$, $\bu^* \in \R^d$ ({cf.}, \eqref{eqn:chapSparse-PACBupperbound-defrho}), we get that
\begin{align*}
\sum_{t=1}^T (y_t - \widehat{y}_t)^2 \leq & \inf_{\bu^* \in \R^d} \left\{\int_{\R^d} \sum_{t=1}^T \bigl(y_t - \bu \cdot \bphi(x_t)\bigr)^2 \rho_{\bu^*,\tau}(\d\bu) + 8 B_{T+1}^2 \KL(\rho_{\bu^*,\tau}, \pi_{\tau})\right\} +  5 B_{T+1}^2 \\
\leq & \inf_{\bu^* \in \R^d} \left\{\sum_{t=1}^T \bigl(y_t - \bu^* \cdot \bphi(x_t)\bigr)^2 + 32 B_{T+1}^2 \norm[\bu^*]_0 \ln\left(1 + \frac{\norm[\bu^*]_1}{\norm[\bu^*]_0 \tau}\right)\right\} \\
& + \tau^2 \sum_{j=1}^d \sum_{t=1}^T \phi_j^2(x_t) + 5 B_{T+1}^2~,
\end{align*}
where the last inequality follows from Lemmas~\ref{lem:chapSparse-PACBupperbound-int} and~\ref{lem:chapSparse-PACBupperbound-KL}.
\end{proofref}

\subsection{A Fully Automatic Algorithm}
\label{sec:chapSparse-onlineSEW-unknownByBphi}

In the previous section, we proved that adaptation to $B_y$ was possible. If we also no longer assume that a bound $B_{\Phi}$ on the trace of the empirical Gram matrix is available to the forecaster, then we can use a doubling trick on the nondecreasing quantity
\[
\gamma_t \eqdef \ln\left(1 + \sqrt{\sum_{s=1}^t \sum_{j=1}^d \phi_j^2(x_s)} \, \right)
\]
and repeatedly run the algorithm $\rm{SeqSEW}^*_{\tau}$ of the previous section for rapidly-decreasing values of~$\tau$. This yields a sparsity regret bound with extra logarithmic multiplicative factors as compared to Proposition~\ref{prop:chapSparse-SRB-unknownBy}, but which holds for a fully automatic algorithm; see Theorem~\ref{thm:chapSparse-SRB-unknownByBphi} below.

More formally, our algorithm $\rm{SeqSEW}^*_*$ is defined as follows. The set of all time rounds $t=1,2, \ldots$ is partitioned into regimes $r = 0, 1, \ldots$ whose final time instances $t_r$ are data-driven. Let $t_{-1} \eqdef 0$ by convention. We call \emph{regime $r$}, $r = 0, 1, \ldots$, the sequence of time rounds $(t_{r-1}+1, \ldots, t_r)$ where $t_r$ is the first date $t \geq t_{r-1}+1$ such that $\gamma_t > 2^r$. At the beginning of regime $r$, we restart the algorithm $\rm{SeqSEW}^*_{\tau}$ defined in Figure~\ref{fig:chapSparse-continuouseEWA-varyingeta-def} with the parameter $\tau$ set to $\tau_r \eqdef 1/\bigl(\exp(2^r)-1\bigr)$.

In particular, on each regime $r$, the current instance of the algorithm $\rm{SeqSEW}^*_{\tau_r}$ only uses the past observations $y_s$, $s \in \{t_{r-1}+1,\ldots,t-1\}$, to perform the online trunction and to tune the inverse temperature parameter. Therefore, the algorithm $\rm{SeqSEW}^*_*$ is fully automatic.

\begin{theorem}
\label{thm:chapSparse-SRB-unknownByBphi}
Without requiring any preliminary knowledge at the beginning of the prediction game, $\rm{SeqSEW}^*_*$ satisfies, for all $T \geq 1$ and all $(x_1, y_1), \ldots, (x_T, y_T) \in \cX \times \R$,
\vspace{-0.1cm}
\begin{align*}
\sum_{t=1}^T (y_t - \widehat{y}_t)^2 \leq \inf_{\bu \in \R^d} & \Biggl\{\sum_{t=1}^T \bigl(y_t - \bu \cdot \bphi(x_t)\bigr)^2 + 128 \Bigl(\max_{1 \leq t \leq T} y_t^2\Bigr) \norm[\bu]_0 \ln\!\left(e + \sqrt{\sum_{t=1}^T \sum_{j=1}^d \phi_j^2(x_t)} \, \right)\\[-0.1cm]
& \quad + \, 32 \Bigl(\max_{1 \leq t \leq T} y_t^2\Bigr) A_T \norm[\bu]_0  \ln\left(1 + \frac{\norm[\bu]_1}{\norm[\bu]_0}\right)\Biggr\} + \, \Bigl(1 + 9 \max_{1 \leq t \leq T} y_t^2\Bigr) A_T~, \nonumber
\end{align*}
where $A_T \eqdef 2 + \log_2 \ln \left(e + \sqrt{\sum_{t=1}^T \sum_{j=1}^d \phi_j^2(x_t)} \, \right)$.
\end{theorem}

Though the algorithm $\rm{SeqSEW}^*_*$ is fully automatic, two possible improvements could be addressed in the future. From a theoretical viewpoint, can we contruct a fully automatic algorithm with a bound similar to Theorem~\ref{thm:chapSparse-SRB-unknownByBphi} but without the extra logarithmic factor $A_T$? From a practical viewpoint, is it possible to perform the adaptation to $B_{\Phi}$ without restarting the algorithm repeatedly (just like we did for $B_y$)? A smoother time-varying tuning $(\tau_t)_{t \geq 2}$ might enable to answer both questions. This would be very probably at the price of a more involved analysis (e.g., if we adapt the PAC-Bayesian bound of Lemma~\ref{lem:chapSparse-PACB-unknownBy}, then a third approximation term would appear in~\eqref{eqn:chapSparse-PACB-unknownBy-spliteta} since $\pi_{\tau_t}$ changes over time).

\begin{proofsketchref}{Theorem~\ref{thm:chapSparse-SRB-unknownByBphi}}
The proof relies on the use of Corollary~\ref{cor:chapSparse-SRB-unknownBy-tauOptimal} on all regimes~$r$ visited up to time $T$. More precisely, note that $\gamma_{t_r-1} \leq 2^r$ by definition of $t_r$ (except maybe in the trivial case when $t_r=t_{r-1}+1$), which entails that
\[
\sum_{t=t_{r-1}+1}^{t_r-1} \sum_{j=1}^d \phi_j^2(x_t) \leq \left(e^{2^r}-1\right)^2 \eqdef B_{\Phi,r}~.
\]
Since we tuned the instance of the algorithm $\rm{SeqSEW}^*_{\tau}$ on regime $r$ with $\tau=\tau_r \eqdef 1/\sqrt{B_{\Phi,r}}$, we can apply Corollary~\ref{cor:chapSparse-SRB-unknownBy-tauOptimal} on regime $r$ for all $r$. Summing the corresponding regret bounds over $r$ then yields the desired result. See Appendix~\ref{sec:chapSparse-proofs-fullyauto} for a detailed proof.
\end{proofsketchref}

Theorem~\ref{thm:chapSparse-SRB-unknownByBphi} yields the following corollary. It upper bounds the regret of the algorithm $\rm{SeqSEW}^*_*$ uniformly over all $\bu \in \R^d$ such that $\norm[\bu]_0 \leq s$ and $\norm[\bu]_1 \leq U$, where the sparsity level $s \in \N$ and the $\ell^1$-diameter $U>0$ are both  unknown to the forecaster. The proof is postponed to Appendix~\ref{sec:chapSparse-proofs-fullyauto}.

\begin{corollary}
\label{cor:chapSparse-SRB-unknownByBphi}
Fix $s \in \N$ and $U > 0$. Then, for all $T \geq 1$ and all $(x_1, y_1), \ldots, (x_T, y_T) \in \cX \times \R$, the regret of the algorithm $\rm{SeqSEW}^*_*$ on $\bigl\{\bu \in \R^d: \norm[\bu]_0 \leq s \; \textrm{and} \; \norm[\bu]_1 \leq U\bigr\}$ is bounded by
\vspace{-0.2cm}
\begin{align*}
& \sum_{t=1}^T (y_t - \widehat{y}_t)^2 - \inf_{\substack{\norm[\bu]_0 \leq s \\ \Arrowvert \bu \Arrowvert_1 \leq U}} \sum_{t=1}^T \bigl(y_t - \bu \cdot \bphi(x_t)\bigr)^2 \\[-0.1cm]
& \qquad \leq 128 \Bigl(\max_{1 \leq t \leq T} y_t^2\Bigr) \, s \, \ln\left(e + \sqrt{\sum_{t=1}^T \sum_{j=1}^d \phi_j^2(x_t)} \, \right) + \, 32 \Bigl(\max_{1 \leq t \leq T} y_t^2\Bigr) A_T \, s \,  \ln\!\left(1 + \frac{U}{s}\right) \nonumber \\
& \qquad \quad + \, \Bigl(1 + 9 \max_{1 \leq t \leq T} y_t^2\Bigr) A_T~, \nonumber
\end{align*}
where $A_T \eqdef 2 + \log_2 \ln \left(e + \sqrt{\sum_{t=1}^T \sum_{j=1}^d \phi_j^2(x_t)} \, \right)$.
\end{corollary}

\section{Adaptivity to the Unknown Variance in the Stochastic Setting}
\label{sec:chapSparse-adaptivityVariance}

In this section, we apply the online algorithm $\rm{SeqSEW}^*_{\tau}$ of Section~\ref{sec:chapSparse-onlineSEW-unknownBy} to two related stochastic settings: the regression model with random design (Section~\ref{sec:chapSparse-adaptivityVariance-random}) and the regression model with fixed design (Section~\ref{sec:chapSparse-adaptivityVariance-fixed}). The sparsity regret bounds proved for this algorithm on individual sequences imply in both settings sparsity oracle inequalities with leading constant~$1$. These risk bounds are of the same flavor as in \citet{DaTsy08SEW, DaTsy10MirrorAveraging} but they are adaptive (up to a logarithmic factor) to the unknown variance~$\sigma^2$ of the noise if the latter is Gaussian. In particular, we solve two questions left open by \citet{DaTsy10MirrorAveraging} in the random design case.

In the sequel, just like in the online deterministic setting, we assume that the forecaster has access to a dictionary $\bphi=(\phi_1,\ldots,\phi_d)$ of measurable base forecasters $\phi_j:\cX \to \R$, $j=1,\ldots,d$.

\subsection{Regression Model With Random Design}
\label{sec:chapSparse-adaptivityVariance-random}

In this section we apply the algorithm $\rm{SeqSEW}^*_{\tau}$ to the regression model with random design. In this batch setting the forecaster is given at the beginning of the game $T$ independent random copies $(X_1,Y_1), \ldots, (X_T,Y_T)$ of $(X,Y) \in \cX \times \R$ whose common distribution is unknown. We assume thereafter that $\E[Y^2]<\infty$; the goal of the forecaster is to estimate the regression function $f:\cX \to \R$ defined by $f(x) \eqdef \E[Y|X=x]$ for all $x \in \cX$. Setting $\epsilon_t \eqdef Y_t - f(X_t)$ for all $t=1,\ldots,T$, note that
\[
Y_t = f(X_t) + \epsilon_t~, \quad 1 \leq t \leq T~,
\]
and that the pairs $(X_1,\epsilon_1),\ldots,(X_T,\epsilon_T)$ are {i.i.d.}\ and such that $\E[\epsilon_1^2]<\infty$ and $\E[\epsilon_1|X_1]=0$ almost surely. In the sequel, we denote the distribution of $X$ by $P^X$ and we set, for all measurable functions $h:\cX \to \R$,
\[
\norm[h]_{L^2} \eqdef \left(\int_{\cX} h(x)^2 P^X(\dd x)\right)^{1/2} = \Bigl(\E\bigl[h(X)^2\bigr]\Bigr)^{1/2}~.
\]
Next we construct an estimator $\widehat{f}_T:\cX \to \R$ based on the sample $(X_1,Y_1),\ldots,(X_T,Y_T)$ that satisfies a sparsity oracle inequality, that is, its expected $L^2$-risk $\E\bigl[\big\Arrowvert f-\widehat{f}_T \big\Arrowvert_{L^2}^2\bigr]$ is almost as small as the smallest $L^2$-risk $\Norm{f-\bu \cdot \bphi}_{L^2}^2$, $\bu \in \R^d$, up to some additive term proportional to $\norm[\bu]_0$.

\subsubsection{Algorithm and Main Result}

Even if the whole sample $(X_1,Y_1), \ldots, (X_T,Y_T)$ is available at the beginning of the prediction game, we treat it in a sequential fashion. We run the algorithm $\rm{SeqSEW}^*_{\tau}$ of Section~\ref{sec:chapSparse-onlineSEW-unknownBy} from time $1$ to time $T$ with $\tau = 1 / \sqrt{d T}$ (note that $T$ is known in this setting). Using the standard online-to-batch conversion (see, e.g., \citealt{Lit-89-Online2Batch,CeCoGen04GeneralizationAbility}), we define our estimator $\widehat{f}_T: \cX \to \R$ as the uniform average
\begin{equation}
\label{eqn:chapSparse-stochastic-random-defAverage1}
\widehat{f}_T \eqdef \frac{1}{T} \sum_{t=1}^T \widetilde{f}_t
\end{equation}
of the estimators $\widetilde{f}_t: \cX \to \R$ sequentially built by the algorithm $\rm{SeqSEW}^*_{\tau}$ as
\begin{equation}
\label{eqn:chapSparse-stochastic-random-defAverage2}
\widetilde{f}_t(x) \eqdef \int_{\R^d} \bigl[\bu \cdot \bphi(x)\bigr]_{B_t} \, p_t(\d\bu)~.
\end{equation}

Note that, contrary to much prior work from the statistics community such as those of \citet{Catoni01StFlour}, \citet{BuNo08SeqProcedures} and \citet{DaTsy10MirrorAveraging}, the estimators $\widetilde{f}_t: \cX \to \R$ are tuned online. Therefore, $\widehat{f}_T$ does not depend on any prior knowledge on the unknown distribution of the $(X_t,Y_t)$, $1 \leq t \leq T$, such as the unknown variance $\E\bigl[(Y-f(X))^2\bigr]$ of the noise, the norms $\norm[\phi_j]_{\infty}$, or the norms $\norm[f - \phi_j]_{\infty}$ (actually, the functions $\phi_j$ and $f - \phi_j$ do not even need to be bounded in $\ell^{\infty}$-norm).

In this respect, this work improves on that of \citet{BuNo08SeqProcedures} who tune their online forecasters as a function of $\norm[f]_{\infty}$ and $\sup_{\bu \in \cU} \norm[\bu \cdot \bphi]_{\infty}$, where $\cU \subset \R^d$ is a bounded comparison set.\footnote{\citet{BuNo08SeqProcedures} study the case where $\cU$ is the (scaled) simplex in $\R^d$ or the set of its vertices.} Their technique is not appropriate when $\norm[f]_{\infty}$ is unknown and it cannot be extended to the case where $\cU = \R^d$ (since $\sup_{\bu \in \R^d} \norm[\bu \cdot \bphi]_{\infty} = + \infty$ if $\bphi \neq {\bf 0}$). The major technical difference is that we truncate the base forecasts $\bu \cdot \bphi(X_t)$ instead of truncating the observations $Y_t$. In particular, this enables us to aggregate the base forecasters $\bu \cdot \bphi$ for all $\bu \in \R^d$, that is, over the whole $\R^d$ space.

The next sparsity oracle inequality is the main result of this section. It follows from the deterministic regret bound of Corollory~\ref{cor:chapSparse-SRB-unknownBy-tauSimple} and from Jensen's inequality. Two corollaries are to be derived later.

\begin{theorem}
\label{thm:chapSparse-stochastic-tuningSimple}
Assume that $(X_1,Y_1), \ldots, (X_T,Y_T) \in \cX \times \R$ are independent random copies of $(X,Y) \in \cX \times \R$, where $\E[Y^2] < +\infty$ and $\norm[\phi_j]_{L^2}^2 \eqdef \E[\phi_j(X)^2] < +\infty$ for all $j=1, \ldots, d$. Then, the estimator $\widehat{f}_T$ defined in \eqref{eqn:chapSparse-stochastic-random-defAverage1}-\eqref{eqn:chapSparse-stochastic-random-defAverage2} satisfies
\begin{align*}
\E\!\left[\norm[f-\widehat{f}_T]_{L^2}^2\right] \leq & \inf_{\bu \in \R^d} \!\left\{\norm[f-\bu \cdot \bphi]_{L^2}^2 + 32 \, \frac{\E\left[\max_{1 \leq t \leq T} Y_t^2\right]}{T} \, \norm[\bu]_0 \ln\!\left(1 + \frac{\sqrt{d T} \norm[\bu]_1}{\norm[\bu]_0}\right)\right\}  \\
& + \frac{1}{d T} \sum_{j=1}^d \norm[\phi_j]_{L^2}^2 + \, 5 \, \frac{\E\left[\max_{1 \leq t \leq T} Y_t^2\right]}{T}~.
\end{align*}
\end{theorem}

\begin{proofsketchref}{Theorem~\ref{thm:chapSparse-stochastic-tuningSimple}}
By Corollary~\ref{cor:chapSparse-SRB-unknownBy-tauSimple} and by definition of $\widetilde{f}_t$ above and $\widehat{y_t} \eqdef \widetilde{f}_t(X_t)$ in Figure~\ref{fig:chapSparse-continuouseEWA-varyingeta-def}, we have, \emph{almost surely},
\begin{align*}
\sum_{t=1}^T (Y_t - \widetilde{f}_t(X_t))^2 \leq & \inf_{\bu \in \R^d} \!\left\{\sum_{t=1}^T \bigl(Y_t - \bu \cdot \bphi(X_t)\bigr)^2 + 32 \!\left(\max_{1 \leq t \leq T} Y_t^2\right) \norm[\bu]_0 \ln\!\left(1 + \frac{\sqrt{d T} \norm[\bu]_1}{\norm[\bu]_0}\right)\!\right\} \\
& + \frac{1}{d T} \sum_{j=1}^d \sum_{t=1}^T \phi_j^2(X_t) + 5 \, \max_{1 \leq t \leq T} Y_t^2~.
\end{align*}
Taking the expectations of both sides and applying Jensen's inequality yields the desired result. For a detailed proof, see Appendix~\ref{sec:chapSparse-proofs-random}.
\end{proofsketchref}

Theorem~\ref{thm:chapSparse-stochastic-tuningSimple} above can be used under several assumptions on the distribution of the output~$Y$. In all cases, it suffices to upper bound the amplitude $\E\bigl[\max_{1 \leq t \leq T} Y_t^2\bigr]$. We present below a general corollary and explain later why our fully automatic procedure $\hat{f}_T$ solves two questions left open by  \citet{DaTsy10MirrorAveraging} (see Corollary~\ref{cor:chapSparse-stochastic-improvementDaTsy} below).

\subsubsection{A General Corollary}

The next sparsity oracle inequality follows from Theorem~\ref{thm:chapSparse-stochastic-tuningSimple} and from the upper bounds on \linebreak[4] $\E\bigl[\max_{1 \leq t \leq T} Y_t^2\bigr]$ entailed by Lemmas~\ref{lem:chapSparse-maxIneq-SG}--\ref{lem:chapSparse-maxIneq-BM} in Appendix~\ref{sec:chapSparse-tools}. The proof is postponed to Appendix~\ref{sec:chapSparse-proofs-random}.

\begin{corollary}
\label{cor:chapSparse-stochastic-tuningSimple}
Assume that $(X_1,Y_1), \ldots, (X_T,Y_T) \in \cX \times \R$ are independent random copies of $(X,Y) \in \cX \times \R$, that $\sup_{1 \leq j \leq d} \norm[\phi_j]_{L^2}^2 < +\infty$, that  $\E|Y|<+\infty$, and that one of the following assumptions holds on  the distribution of $\Delta Y \eqdef Y - \E[Y]$.
\begin{itemize}
\item {\bf $\bigl(\rm{BD}(B)\bigr)\!:$} $|\Delta Y| \leq B$ almost surely for a given constant $B > 0$;
\item {\bf $\bigl(\rm{SG}(\sigma^2)\bigr)\!:$} $\Delta Y$ is subgaussian with variance factor $\sigma^2>0$, that is, $\E\left[e^{\lambda \Delta Y}\right] \leq e^{\lambda^2 \sigma^2 / 2}$ for all $ \lambda \in \R$;
\item {\bf $\bigl(\rm{BEM}(\alpha,M)\bigr)\!:$} $\Delta Y$ has a bounded exponential moment, that is, $\E\left[e^{\alpha |\Delta Y|}\right] \leq M$ for some given constants $\alpha>0$ and $M>0$;
\item {\bf $\bigl(\rm{BM}(\alpha,M)\bigr)\!:$} $\Delta Y$ has a bounded moment, that is, $\E\bigl[|\Delta Y|^{\alpha}\bigr] \leq M$ for some given constants $\alpha>2$ and $M>0$.
\end{itemize}
Then, the estimator $\widehat{f}_T$ defined above satisfies
\begin{align*}
\E\!\left[\norm[f-\widehat{f}_T]_{L^2}^2\right] \leq & \inf_{\bu \in \R^d} \left\{\norm[f-\bu \cdot \bphi]_{L^2}^2 + 64 \left(\frac{\E[Y]^2}{T} + \psi_T\right) \norm[\bu]_0 \ln\left(1 + \frac{\sqrt{d T} \norm[\bu]_1}{\norm[\bu]_0}\right)\right\}  \\
& + \frac{1}{d T} \sum_{j=1}^d \norm[\phi_j]_{L^2}^2 + \, 10 \left(\frac{\E[Y]^2}{T} + \psi_T\right)~,
\end{align*}
where
\begin{numcases}{\psi_T \eqdef \frac{1}{T} \E\!\left[\max_{1 \leq t \leq T} \bigl(Y_t - \E[Y_t]\bigr)^2\right] \leq}
\nonumber
\frac{B^2}{T} & under Assumption $\bigl(\rm{BD}(B)\bigr)$, \\
\nonumber
\frac{2 \sigma^2 \ln(2 e T)}{T}\!\!\!  & under Assumption $\bigl(\rm{SG}(\sigma^2)\bigr)$, \\
\nonumber
\frac{\ln^2\bigl((M+e) T\bigr)}{\alpha^2 \, T} & under Assumption $\bigl(\rm{BEM}(\alpha,M)\bigr)$, \\
\nonumber
\frac{M^{2/\alpha}}{T^{(\alpha-2)/\alpha}} & under Assumption $\bigl(\rm{BM}(\alpha,M)\bigr)$.
\end{numcases}
\end{corollary}

\vspace{0.3cm}
Several comments can be made about Corollary~\ref{cor:chapSparse-stochastic-tuningSimple}. We first stress that, if $T \geq 2$, then the two ``bias'' terms $\E[Y]^2 / T$ above can be avoided, at least at the price of a multiplicative factor of $2T/(T-1)\leq 4$. This can be achieved via a slightly more sophisticated online clipping---see Remark~\ref{rmk:chapSparse-stochastic-EYnotNecessary} in Appendix~\ref{sec:chapSparse-proofs-random}.

Second, under the assumptions $\bigl(\rm{BD}(B)\bigr)$, $\bigl(\rm{SG}(\sigma^2)\bigr)$, or $\bigl(\rm{BEM}(\alpha,M)\bigr)$, the key quantity $\psi_T$ is respectively of the order of $1/T$, $\ln(T)/T$ and $\ln^2(T)/T$. Up to a logarithmic factor, this corresponds to the classical fast rate of convergence $1/T$ obtained in the random design setting for different aggregation problems (see, e.g., \citealt{Cat-99-UniversalAggregation,JuRiTsy08MirrorAveraging,Au09FastRates} for model-selection-type aggregation and \citealt{DaTsy10MirrorAveraging} for linear aggregation). We were able to get similar rates---with, however, a fully automatic procedure---since our online algorithm $\rm{SeqSEW}^*_{\tau}$ is well suited for bounded individual sequences with an unknown bound. More precisely, the finite {i.i.d.}\ sequence $Y_1,\ldots,Y_T$ is almost surely uniformly bounded by the random bound $\max_{1 \leq t \leq T} |Y_t|$. Our individual sequence techniques adapt sequentially to this random bound, yielding a regret bound that scales as $\max_{1 \leq t \leq T} Y_t^2$. As a result, the risk bounds obtained after the online-to-batch conversion scale as $\E\bigl[\max_{1 \leq t \leq T} Y_t^2\bigr]/T$. If the distribution of the output $Y$ is sufficiently lightly-tailed---which includes the quite general bounded-exponential-moment assumption---then we can recover the fast rate of convergence $1/T$ up to a logarithmic factor.

We note that there is still a question left open for heavy-tailed output distributions. For example, under the bounded moment assumption $\bigl(\rm{BM}(\alpha,M)\bigr)$, the rate $T^{-(\alpha-2)/\alpha}$ that we proved does not match the faster rate $T^{-\alpha/(\alpha+2)}$ obtained by \citet{JuRiTsy08MirrorAveraging} and \citet{Au09FastRates} under a similar assumption. Their methods use some preliminary knowledge on the output distribution (such as the exponent $\alpha$). Thus, obtaining the same rate with a procedure tuned in an automatic fashion---just like our method $\hat{f}_T$---is a challenging task. For this purpose, a different tuning of $\eta_t$ or a more sophisticated online truncation might be necessary.

Third, several variations on the assumptions are possible. First note that several classical assumptions on $Y$ expressed in terms of $f(X)$ and $\epsilon \eqdef Y-f(X)$ are either particular cases of the above corollary or can be treated similarly. Indeed, each of the four assumptions above on $\Delta Y \eqdef Y - \E[Y] = f(X)-\E[f(X)] + \epsilon$ is satisfied as soon as both the distribution of $f(X)-\E[f(X)]$ and the conditional distribution of $\epsilon$ (conditionally on $X$) satisfy the same type of assumption. For example, if $f(X)-\E[f(X)]$ is subgaussian with variance factor $\sigma_X^2$ and if $\epsilon$ is subgaussian conditionally on $X$ with a variance factor uniformly bounded by a constant $\sigma_{\epsilon}^2$, then $\Delta Y$ is subgaussian with variance factor $\sigma_X^2+\sigma_{\epsilon}^2$ (see also Remark~\ref{rmk:chapSparse-stochastic-assump-decoupling} in Appendix~\ref{sec:chapSparse-proofs-random} to avoid conditioning).

The assumptions on $f(X)-\E[f(X)]$ and $\epsilon$ can also be mixed together. For instance, as explained in Remark~\ref{rmk:chapSparse-stochastic-assump-decoupling} in Appendix~\ref{sec:chapSparse-proofs-random}, under the classical assumptions
\begin{equation}
\label{rmk:chapSparse-stochastic-assump1}
\norm[f]_{\infty} < +\infty \qquad \textrm{and} \qquad \E\Bigl[e^{\alpha |\epsilon|}~\Big|~X\Bigr] \leq M \quad \textrm{a.s.}
\end{equation}
or
\begin{equation}
\label{rmk:chapSparse-stochastic-assump2}
\norm[f]_{\infty} < +\infty \qquad \textrm{and} \qquad \E\!\left[e^{\lambda \epsilon}~\Big|~X\right] \leq e^{\lambda^2 \sigma^2 / 2} \quad \textrm{a.s.}, \quad \forall \lambda \in \R~,
\end{equation}
the key quantity $\psi_T$ in the corollary can be bounded from above by
\begin{numcases}{\psi_T \leq}
\nonumber
\frac{8 \norm[f]_{\infty}^2}{T} + \frac{2 \ln^2\bigl((M+e) T\bigr)}{\alpha^2 \, T} & under the set of assumptions~(\ref{rmk:chapSparse-stochastic-assump1}), \\
\nonumber
\frac{8 \norm[f]_{\infty}^2}{T} + \frac{4 \sigma^2 \ln(2 e T)}{T} & under the set of assumptions~(\ref{rmk:chapSparse-stochastic-assump2}).
\end{numcases}

In particular, under the set of assumptions~(\ref{rmk:chapSparse-stochastic-assump2}), our procedure $\hat{f}_T$ solves two questions left open by \citet{DaTsy10MirrorAveraging}. We discuss below our contributions in this particular case.

\subsubsection{Questions Left Open by Dalalyan and Tsybakov}

In this subsection we focus on the case when the set of assumptions~\eqref{rmk:chapSparse-stochastic-assump2} holds true. Namely, the regression function $f$ is bounded (by an unknown constant) and the noise $\epsilon \eqdef Y-f(X)$ is subgaussian conditionally on $X$ with an unknown variance factor $\sigma^2 > 0$. An important particular case is when $\norm[f]_{\infty} < +\infty$ and when the noise $\epsilon$ is independent of $X$ and normally distributed $\cN(0,\sigma^2)$.

Under the set of assumptions \eqref{rmk:chapSparse-stochastic-assump2}, the two terms $\E\bigl[\max_{1 \leq t \leq T} Y_t^2\bigr]$ of Theorem~\ref{thm:chapSparse-stochastic-tuningSimple}  can be upper bounded in a simpler and slightly tighter way as compared to the proof of Corollary~\ref{cor:chapSparse-stochastic-tuningSimple} (we only use the inequality $(x+y)^2 \leq 2 x^2 + 2 y^2$ once, instead of twice). It yields the following sparsity oracle inequality.

\begin{corollary}
\label{cor:chapSparse-stochastic-improvementDaTsy}
Assume that $(X_1,Y_1), \ldots, (X_T,Y_T) \in \cX \times \R$ are independent random copies of $(X,Y) \in \cX \times \R$ such that the set of assumptions~\eqref{rmk:chapSparse-stochastic-assump2} above holds true. Then, the estimator $\widehat{f}_T$ defined in \eqref{eqn:chapSparse-stochastic-random-defAverage1}-\eqref{eqn:chapSparse-stochastic-random-defAverage2} satisfies
\begin{align*}
& \E\!\left[\norm[f-\widehat{f}_T]_{L^2}^2\right] \\
& \qquad \leq \inf_{\bu \in \R^d} \!\left\{\norm[f-\bu \cdot \bphi]_{L^2}^2 + 64 \!\left(\norm[f]_{\infty}^2 + 2 \sigma^2 \ln (2 e T)\right) \frac{\norm[\bu]_0}{T} \ln\!\left(1 + \frac{\sqrt{d T} \norm[\bu]_1}{\norm[\bu]_0}\right)\!\right\} \\
& \qquad \qquad + \frac{1}{d T} \sum_{j=1}^d \norm[\phi_j]_{L^2}^2 + \frac{10}{T} \left(\norm[f]_{\infty}^2 + 2 \sigma^2 \ln (2 e T)\right)~.
\end{align*}
\end{corollary}

\begin{proof}
We apply Theorem~\ref{thm:chapSparse-stochastic-tuningSimple} and bound $\E\bigl[\max_{1 \leq t \leq T} Y_t^2\bigr]$ from above. By the elementary inequality $(x+y)^2 \leq 2 x^2 + 2 y^2$ for all $x,y \in \R$, we get
\vspace{-0.1cm}
\begin{align*}
\E\!\left[\max_{1 \leq t \leq T} Y_t^2\right] & = \E\!\left[\max_{1 \leq t \leq T} \bigl( f(X_t) + \epsilon_t \bigr)^2\right] \leq 2 \left(\norm[f]_{\infty}^2 + \E\!\left[\max_{1 \leq t \leq T} \epsilon_t^2\right] \right) \\
& \leq 2 \left(\norm[f]_{\infty}^2 + 2 \sigma^2 \ln (2 e T)\right)~,
\end{align*}
where the last inequality follows from Lemma~\ref{lem:chapSparse-maxIneq-SG} in Appendix~\ref{sec:chapSparse-tools} and from the fact that, for all $1 \leq t \leq T$ and all $\lambda \in \R$, we have $\E\!\left[e^{\lambda \epsilon_t} \right] = \E\!\left[e^{\lambda \epsilon} \right] = \E\!\left[ \E\!\left[e^{\lambda \epsilon} \, \big| \, X \right] \right] \leq e^{\lambda^2 \sigma^2 / 2}$ by \eqref{rmk:chapSparse-stochastic-assump2}. (Note that the assumption of conditional subgaussianity in \eqref{rmk:chapSparse-stochastic-assump2} is stronger than what we need, that is, subgaussianity without conditioning.) This concludes the proof.
\end{proof}

The above bound is of the same order (up to a $\ln T$ factor) as the sparsity oracle inequality proved in Proposition~1 of \citet{DaTsy10MirrorAveraging}. For the sake of comparison we state below with our notations (e.g., $\beta$ therein corresponds to $1 / \eta$ in this paper) a straightforward consequence of this proposition, which follows by Jensen's inequality and the particular\footnote{Proposition~1 of~\citet{DaTsy10MirrorAveraging} may seem more general than Corollary~\ref{cor:chapSparse-stochastic-improvementDaTsy} at first sight since it holds for all $\tau > 0$, but this is actually also the case for Corollary~\ref{cor:chapSparse-stochastic-improvementDaTsy}. The proof of the latter would indeed have remained true had we replaced $\tau=1/\sqrt{d T}$ with any value of $\tau > 0$ (see~Proposition~\ref{prop:chapSparse-SRB-unknownBy}). We however chose the reasonable value $\tau=1/\sqrt{d T}$ to make our algorithm parameter-free. As noted earlier, if $\norm[\bphi]_{\infty} \eqdef \sup_{x \in \cX} \max_{1 \leq j \leq d} |\phi_j(x)|$ is finite and known by the forecaster, another simple and easy-to-analyse tuning is given by $\tau = 1/(\norm[\bphi]_{\infty} \!\sqrt{d \, T})$.} choice $\tau=\min\bigl\{1/\sqrt{d T},R/(4d) \bigr\}$.

\begin{proposition}[A consequence of Prop.~1 of \citealt{DaTsy10MirrorAveraging}]
\label{prop:chapSparse-DaTsyRandomDesign}
\ \\
Assume that $\sup_{1 \leq j \leq d} \norm[\phi_j]_{\infty} < \infty$ and that the set of assumptions~(\ref{rmk:chapSparse-stochastic-assump2}) above holds true. Then, for all $R > 0$ and all $\eta \leq \bar{\eta}(R) \eqdef \bigl(2 \sigma^2 + 2 \sup_{\norm[\bu]_1 \leq R} \norm[\bu \cdot \bphi - f]^2_{\infty}\bigr)^{-1}$, the mirror averaging aggregate $\widehat{f}_T:\cX \to \R$ defined by \citet[Equations~(1) and~(3)]{DaTsy10MirrorAveraging} with $\tau=\min\bigl\{1/\sqrt{d T},R/(4d) \bigr\}$ satisfies
\begin{align*}
\E\!\left[\norm[f-\widehat{f}_T]_{L^2}^2\right] & \leq \inf_{\norm[\bu]_1 \leq R/2} \Biggl\{\norm[f-\bu \cdot \bphi]_{L^2}^2 + \frac{4}{\eta} \, \frac{\norm[\bu]_0}{T+1} \ln\!\left(1+\frac{\sqrt{d T} \norm[\bu]_1+2d}{\norm[\bu]_0}\right)\Biggr\} \\
& \qquad + \frac{4}{d T} \sum_{j=1}^d \norm[\phi_j]_{L^2}^2 + \frac{1}{(T+1) \eta}~.
\end{align*}
\end{proposition}

\label{text:DaTsy-openQuestions}
We can now discuss the two questions left open by \citet{DaTsy10MirrorAveraging}.

\noindent{\em Risk bound on the whole $\R^d$ space.} Despite the similarity of the two bounds, the sparsity oracle inequality stated in Proposition~\ref{prop:chapSparse-DaTsyRandomDesign} above only holds for vectors $\bu$ within an $\ell^1$-ball of finite radius $R/2$, while our bound holds over the whole $\R^d$ space. Moreover, the parameter $R$ above has to be chosen in advance, but it cannot be chosen too large since $1/\eta \geq 1/\bar{\eta}(R)$, which grows as $R^2$ when $R \to +\infty$ (if $\bphi \neq {\bf 0}$). \citet[Section~4.2]{DaTsy10MirrorAveraging} thus asked whether it was possible to get a bound with $1/\eta < + \infty$ such that the infimum in Proposition~\ref{prop:chapSparse-DaTsyRandomDesign} extends to the whole $\R^d$ space. Our results show that, thanks to data-driven truncation, the answer is positive.

Note that it is still possible to transform the bound of Proposition~\ref{prop:chapSparse-DaTsyRandomDesign} into a bound over the whole $\R^d$ space if the parameter $R$ is chosen (illegally) as $R = 2 \norm[\bu^*]_1$ (or as a tight upper bound of the last quantity), where $\bu^* \in \R^d$ minimizes over $\R^d$ the regularized risk
\begin{align*}
\norm[f-\bu \cdot \bphi]_{L^2}^2 & + \frac{4}{\bar{\eta}(2 \norm[\bu]_1)} \, \frac{\norm[\bu]_0}{T+1} \ln\!\left(1+\frac{\sqrt{d T} \norm[\bu]_1 +2d}{\norm[\bu]_0}\right) \\
& + \frac{4}{d T} \sum_{j=1}^d \norm[\phi_j]_{L^2}^2 + \frac{1}{(T+1) \, \bar{\eta}(2\norm[\bu]_1)} ~.
\end{align*}
For instance, choosing $R = 2 \norm[\bu^*]_1$ and $\eta = \bar{\eta}(R)$, we get from Proposition~\ref{prop:chapSparse-DaTsyRandomDesign} that the expected $L^2$-risk $\E\bigl[\Arrowvert f-\widehat{f}_T\Arrowvert_{L^2}^2\bigr]$ of the corresponding procedure is upper bounded by the infimum of the above regularized risk over all $\bu \in \R^d$. However, this parameter tuning is illegal since $\norm[\bu^*]_1$ is not known in practice. On the contrary, thanks to data-driven truncation, the prior knowledge of $\norm[\bu^*]_1$ is not required by our procedure.

\noindent{\em Adaptivity to the unknown variance of the noise.} The second open question, which was raised by \citet[Section~5.1, Remark~6]{DaTsy10MirrorAveraging}, deals with the prior knowledge of the variance factor $\sigma^2$ of the noise. The latter is indeed required by their algorithm for the choice of the inverse temperature parameter $\eta$. Since the noise level $\sigma^2$ is unknown in practice, the authors asked the important question whether adaptivity to $\sigma^2$ was possible. Up to a $\ln T$ factor, Corollary~\ref{cor:chapSparse-stochastic-improvementDaTsy} above provides a positive answer.

\subsection{Regression Model With Fixed Design}
\label{sec:chapSparse-adaptivityVariance-fixed}

In this section, we consider the regression model with fixed design. In this batch setting the forecaster is given at the beginning of the game a $T$-sample $(x_1,Y_1), \ldots, (x_T,Y_T) \in \cX \times \R$, where the $x_t$ are deterministic elements in $\cX$ and where
\begin{equation}
\label{eqn:chapSparse-stochastic-fixedDesign-setting}
Y_t = f(x_t) + \epsilon_t~, \quad 1 \leq t \leq T,
\end{equation}
for some {i.i.d.}\ sequence $\epsilon_1,\ldots,\epsilon_T \in \R$ (with unknown distribution) and some unknown function $f:\cX \rightarrow \R$. Next we construct an estimator $\widehat{f}_T:\cX \to \R$ of $f$ based on the sample $(x_1,Y_1),\ldots,(x_T,Y_T)$ that satisfies a sparsity oracle inequality, that is, its expected mean squared error $\E\bigl[\frac{1}{T} \sum_{t=1}^T (f(x_t)-\widehat{f}_T(x_t))^2\bigr]$ is almost as small as the smallest mean squared error $\frac{1}{T} \sum_{t=1}^T (f(x_t)-\bu \cdot \bphi(x_t))^2$, $\bu \in \R^d$, up to some additive term proportional to $\norm[\bu]_0$.

In this setting, just like in Section~\ref{sec:chapSparse-adaptivityVariance-random}, our algorithm and the corresponding analysis are a straightforward consequence of the general results on individual sequences developed in Section~\ref{sec:chapSparse-sparseOnlinePrediction-deterministic}. As in the random design setting, the sample $(x_1,Y_1), \ldots, (x_T,Y_T)$ is treated in a sequential fashion. We run the algorithm $\rm{SeqSEW}^*_{\tau}$ defined in Figure~\ref{fig:chapSparse-continuouseEWA-varyingeta-def} from time $1$ to time $T$ with the particular choice of $\tau = 1 / \sqrt{d T}$. We then define our estimator $\widehat{f}_T:\cX \to \R$ by
\begin{equation}
\label{eqn:chapSparse-stochastic-fixedDesign-algo1}
\widehat{f}_T(x) \eqdef \left\{ \begin{array}{ll}
	 \displaystyle{\frac{1}{n_x} \sum_{\substack{1 \leq t \leq T\\ t:x_t=x}} \widetilde{f}_t(x)} & \textrm{if $x \in \{x_1, \ldots, x_T\}$~,} \\
	0 & \textrm{if $x \notin \{x_1, \ldots, x_T\}$~,}
\end{array} \right.
\end{equation}
where $n_x \eqdef \bigl|\bigl\{t:x_t=x\bigr\}\bigr| = \sum_{t=1}^T \indic{x_t=x}$, and
where the estimators $\widetilde{f}_t: \cX \rightarrow \R$ sequentially built by the algorithm $\rm{SeqSEW}^*_{\tau}$ are defined by
\begin{equation}
\label{eqn:chapSparse-stochastic-fixedDesign-algo3}
\widetilde{f}_t(x) \eqdef \int_{\R^d} \bigl[\bu \cdot \bphi(x)\bigr]_{B_t} \, p_t(\d\bu)~.
\end{equation}
In the particular case when the $x_t$ are all distinct, $\widehat{f}_T$ is simply defined by $\widehat{f}_T(x_t) \eqdef \widetilde{f}_t(x_t)$ for all $t \in \{1, \ldots, T\}$ and by $\widehat{f}_T(x) = 0$ otherwise. Therefore, in this case, $\hat{f}_T$ only uses the observations $y_1,\ldots,y_{t-1}$ to estimate $f(x_t)$ (in particular, $\hat{f}_T(x_1)$ is deterministic).\\

The next theorem is the main result of this subsection. It follows as in the random design setting from the deterministic regret bound of Corollory~\ref{cor:chapSparse-SRB-unknownBy-tauSimple} and from Jensen's inequality. The proof is postponed to Appendix~\ref{sec:chapSparse-proofs-fixed}.

\begin{theorem}
\label{thm:chapSparse-stochastic-fixedDesign-tuningSimple}
Consider the regression model with fixed design described in \eqref{eqn:chapSparse-stochastic-fixedDesign-setting}. Then, the estimator $\widehat{f}_T$ defined in \eqref{eqn:chapSparse-stochastic-fixedDesign-algo1}--\eqref{eqn:chapSparse-stochastic-fixedDesign-algo3} satisfies
\begin{align*}
\E\!\left[\frac{1}{T} \sum_{t=1}^T \bigl(f(x_t)-\widehat{f}_T(x_t)\bigr)^2\right] \leq & \inf_{\bu \in \R^d} \Biggl\{\frac{1}{T} \sum_{t=1}^T \bigl(f(x_t)-\bu \cdot \bphi(x_t)\bigr)^2 \\
& + 32 \, \frac{\E\left[\max_{1 \leq t \leq T} Y_t^2\right]}{T} \, \norm[\bu]_0 \, \ln\left(1 + \frac{\sqrt{d T} \, \norm[\bu]_1}{\norm[\bu]_0}\right)\Biggr\}  \\
& + \frac{1}{d T^2} \sum_{j=1}^d \sum_{t=1}^T \phi_j^2(x_t) + \, 5 \, \frac{\E\left[\max_{1 \leq t \leq T} Y_t^2\right]}{T}~.
\end{align*}
\end{theorem}

As in Section~\ref{sec:chapSparse-adaptivityVariance-random}, the amplitude $\E\bigl[\max_{1 \leq t \leq T} Y_t^2\bigr]$ can be upper bounded under various assumptions. The proof of the following corollary is postponed to Appendix~\ref{sec:chapSparse-proofs-fixed}.

\begin{corollary}
\label{cor:chapSparse-stochastic-fixedDesign-tuningSimple}
Consider the regression model with fixed design described in \eqref{eqn:chapSparse-stochastic-fixedDesign-setting}. Assume that one of the following assumptions holds on the distribution of $\epsilon_1$.
\begin{itemize}
\item {\bf $\bigl(\rm{BD}(B)\bigr)\!:$} $|\epsilon_1| \leq B$ almost surely for a given constant $B > 0$;
\item {\bf $\bigl(\rm{SG}(\sigma^2)\bigr)\!:$} $\epsilon_1$ is subgaussian with variance factor $\sigma^2>0$, that is, $\E\left[e^{\lambda \epsilon_1}\right] \leq e^{\lambda^2 \sigma^2 / 2}$ for all $ \lambda \in \R$;
\item {\bf $\bigl(\rm{BEM}(\alpha,M)\bigr)\!:$} $\epsilon$ has a bounded exponential moment, that is, $\E\left[e^{\alpha |\epsilon|}\right] \leq M$ for some given constants $\alpha>0$ and $M>0$;
\item {\bf $\bigl(\rm{BM}(\alpha,M)\bigr)\!:$} $\epsilon$ has a bounded moment, that is, $\E\bigl[|\epsilon|^{\alpha}\bigr] \leq M$ for some given constants $\alpha>2$ and $M>0$.
\end{itemize}
Then, the estimator $\widehat{f}_T$ defined in \eqref{eqn:chapSparse-stochastic-fixedDesign-algo1}--\eqref{eqn:chapSparse-stochastic-fixedDesign-algo3} satisfies
\begin{align*}
\E\!\left[\frac{1}{T} \sum_{t=1}^T \bigl(f(x_t)-\widehat{f}_T(x_t)\bigr)^2\right] \leq & \inf_{\bu \in \R^d} \Biggl\{\frac{1}{T} \sum_{t=1}^T \bigl(f(x_t)-\bu \cdot \bphi(x_t)\bigr)^2 \\
& + 64 \left(\frac{\max_{1 \leq t \leq T} f^2(x_t)}{T} + \psi_T\right) \norm[\bu]_0 \, \ln\left(1 + \frac{\sqrt{d T} \, \norm[\bu]_1}{\norm[\bu]_0}\right)\Biggr\}  \\
& + \frac{1}{d T^2} \sum_{j=1}^d \sum_{t=1}^T \phi_j^2(x_t) + \, 10 \left(\frac{\max_{1 \leq t \leq T} f^2(x_t)}{T} + \psi_T\right)~,
\end{align*}
where
\begin{numcases}{\psi_T \eqdef \frac{1}{T} \, \E\!\left[\max_{1 \leq t \leq T} \epsilon_t^2\right] \leq}
\nonumber
\frac{B^2}{T} & if Assumption $\bigl(\rm{BD}(B)\bigr)$ holds, \\
\nonumber
\frac{2 \sigma^2 \ln(2 e T)}{T} & if Assumption $\bigl(\rm{SG}(\sigma^2)\bigr)$ holds, \\
\nonumber
\frac{\ln^2\left((M+e) T\right)}{\alpha^2 \, T} & if Assumption $\bigl(\rm{BEM}(\alpha,M)\bigr)$ holds, \\
\nonumber
\frac{M^{2/\alpha}}{T^{(\alpha-2)/\alpha}} & if Assumption $\bigl(\rm{BM}(\alpha,M)\bigr)$ holds.
\end{numcases}
\end{corollary}

The above bound is of the same flavor as that of \citet[Theorem~5]{DaTsy08SEW}. It has one advantage and one drawback. On the one hand, we note two additional ``bias'' terms $\bigl(\max_{1 \leq t \leq T} f^2(x_t)\bigr)/T$ as compared to the bound of \citet[Theorem~5]{DaTsy08SEW}. As of now, we have not been able to remove them using ideas similar to what we did in the random design case (see Remark~\ref{rmk:chapSparse-stochastic-EYnotNecessary} in Appendix~\ref{sec:chapSparse-proofs-random}). On the other hand, under Assumption $\bigl(\rm{SG}(\sigma^2)\bigr)$, contrary to \citet{DaTsy08SEW}, our algorithm does not require the prior knowledge of the variance factor $\sigma^2$ of the noise.

\acks{The author would like to thank Arnak Dalalyan, Gilles Stoltz, and Pascal Massart for their helpful feedback and suggestions, as well as two anonymous reviewers for their insightful comments, one of which helped us simplify the online tuning carried out in Section~\ref{sec:chapSparse-onlineSEW-unknownBy}. The author acknowledges the support of the French Agence Nationale de la Recherche (ANR), under grant PARCIMONIE (\url{http://www.proba.jussieu.fr/ANR/Parcimonie}), and of the IST Programme of the European Community, under the PASCAL2 Network of Excellence, IST-2007-216886.}

\appendix

\section{Proofs}
\label{sec:chapSparse-proofs}

In this appendix we provide the proofs of some results stated above.

\subsection{Proofs of Theorem~\ref{thm:chapSparse-SRB-unknownByBphi} and Corollary~\ref{cor:chapSparse-SRB-unknownByBphi}}
\label{sec:chapSparse-proofs-fullyauto}

Before proving Theorem~\ref{thm:chapSparse-SRB-unknownByBphi}, we first need the following comment. Since the algorithm $\rm{SeqSEW}^*_{\tau}$ is restarted at the beginning of each regime, the threshold values $B_t$ used on regime $r$ by the algorithm $\rm{SeqSEW}^*_{\tau}$  are not computed on the basis of all past observations $y_1,\ldots,y_{t-1}$ but only on the basis of the past observations $y_t$, $t \in \{t_{r-1}+1,\ldots,t-1\}$. To avoid any ambiguity, we set $B_{r,t_{r-1}+1} \eqdef 0$ and
\begin{equation*}
B_{r,t} \eqdef \max_{t_{r-1}+1 \leq s \leq t-1} |y_s|~, \quad t \in \{t_{r-1}+2, \ldots, t_r\}~.
\end{equation*}

\begin{proofref}{Theorem~\ref{thm:chapSparse-SRB-unknownByBphi}}
We denote by $R \eqdef \min\{r \in \N: T \leq t_r\}$ the index of the last regime. For notational convenience, we re-define $t_R \eqdef T$ (even if $\gamma_T \leq 2^R$). \\

We upper bound the regret of the algorithm $\rm{SeqSEW}^*_*$ on $\{1,\ldots,T\}$ by the sum of its regrets on each time interval. To do so, first note that\footnote{In the trivial cases where $t_r=t_{r-1}+1$ for some $r$, the sum $\sum_{t=t_{r-1}+1}^{t_r-1} (y_t - \widehat{y}_t)^2$ equals $0$ by convention.}
\begin{align}
\sum_{t=1}^T (y_t - \widehat{y}_t)^2 = & \sum_{r=0}^R \sum_{t=t_{r-1}+1}^{t_r} (y_t - \widehat{y}_t)^2  =  \sum_{r=0}^R \left((y_{t_r} - \widehat{y}_{t_r})^2 + \sum_{t=t_{r-1}+1}^{t_r-1} (y_t - \widehat{y}_t)^2\right)  \nonumber  \\
\leq & \sum_{r=0}^R \left(2 (y_{t_r}^2 + B_{r, t_r}^2) + \sum_{t=t_{r-1}+1}^{t_r-1} (y_t - \widehat{y}_t)^2\right)  \label{eqn:chapSparse-unknownByBphi-SRB-eachPeriod0-beforelast}  \\
\leq & \sum_{r=0}^R \left(\sum_{t=t_{r-1}+1}^{t_r-1} (y_t - \widehat{y}_t)^2\right) + 4 (R+1) {y_T^*}^2~, \label{eqn:chapSparse-unknownByBphi-SRB-eachPeriod0}
\end{align}
where we set $y_T^* \eqdef \max_{1 \leq t \leq T} |y_t|$, where~(\ref{eqn:chapSparse-unknownByBphi-SRB-eachPeriod0-beforelast}) follows from the upper bound $(y_{t_r} - \widehat{y}_{t_r})^2 \leq 2 (y_{t_r}^2+{\widehat{y}_{t_r}}^2) \leq 2 (y_{t_r}^2 + B_{r, t_r}^2)$ (since $|\widehat{y}_{t_r}| \leq B_{r, t_r}$ by construction), and where~(\ref{eqn:chapSparse-unknownByBphi-SRB-eachPeriod0}) follows from the inequalities $y_{t_r}^2 \leq {y_T^*}^2$ and
\[
B_{r, t_r}^2 \eqdef \max_{t_{r-1}+1 \leq t \leq t_r-1} y_t^2 \leq {y_T^*}^2~.
\]
But, for every $r = 0,\ldots,R$, the trace of the empirical Gram matrix on $\{t_{r-1}+1,\ldots,t_r-1\}$ is upper bounded by
\[
\sum_{t=t_{r-1}+1}^{t_r-1} \sum_{j=1}^d \phi_j^2(x_t) \leq \sum_{t=1}^{t_r-1} \sum_{j=1}^d \phi_j^2(x_t) \leq (e^{2^r}-1)^2~,
\]
where the last inequality follows from the fact that $\gamma_{t_r-1} \leq 2^r$ (by definition of $t_r$). Since in addition $\tau_r \eqdef 1/\sqrt{(e^{2^r}-1)^2}$, we can apply Corollory~\ref{cor:chapSparse-SRB-unknownBy-tauOptimal} on each period $\{t_{r-1}+1, \ldots, t_r - 1\}$, $r = 0, \ldots, R$, with $B_{\Phi} = (e^{2^r}-1)^2$ and get from (\ref{eqn:chapSparse-unknownByBphi-SRB-eachPeriod0}) the upper bound
\begin{equation}
\label{eqn:chapSparse-unknownByBphi-SRB-eachPeriod1}
\sum_{t=1}^{T} (y_t - \widehat{y}_t)^2 \leq \sum_{r=0}^R \inf_{\bu \in \R^d} \left\{\sum_{t=t_{r-1}+1}^{t_r-1} \bigl(y_t - \bu \cdot \bphi(x_t)\bigr)^2 + \Delta_r(\bu) \right\} \, + 4 (R+1) {y_T^*}^2~,
\end{equation}
where
\begin{equation}
\label{eqn:chapSparse-unknownByBphi-SRB-eachPeriod1-DeltaBound}
\Delta_r(\bu) \eqdef 32 B_{r, t_r}^2 \norm[\bu]_0 \ln\left(1 + \frac{\bigl(e^{2^r}-1\bigr) \norm[\bu]_1}{\norm[\bu]_0}\right) + 5 B_{r, t_r}^2 + 1~.
\end{equation}
Since the infimum is superadditive and since $\bigl(y_{t_r} - \bu \cdot \bphi(x_{t_r})\bigr)^2 \geq 0$ for all $r=0,\ldots,R$, we get from~\eqref{eqn:chapSparse-unknownByBphi-SRB-eachPeriod1} that
\begin{align}
\sum_{t=1}^{T} (y_t - \widehat{y}_t)^2 \leq & \inf_{\bu \in \R^d} \sum_{r=0}^R \left(\sum_{t=t_{r-1}+1}^{t_r} \bigl(y_t - \bu \cdot \bphi(x_t)\bigr)^2 + \Delta_r(\bu)\right) \, + 4 (R+1) {y_T^*}^2 \nonumber \\
= & \inf_{\bu \in \R^d} \Biggl\{\sum_{t=1}^T \bigl(y_t - \bu \cdot \bphi(x_t)\bigr)^2 + \sum_{r=0}^R \Delta_r(\bu)\Biggr\} \, + 4 (R+1) {y_T^*}^2~. \label{eqn:chapSparse-unknownByBphi-SRB-eachPeriod2}
\end{align}

Let $\bu \in \R^d$. Next we bound $\sum_{r=0}^R \Delta_r(\bu)$ and $4 (R+1) {y_T^*}^2$ from above. First note that, by the upper bound $B_{r, t_r}^2 \leq {y_T^*}^2$ and by the elementary inequality $\ln(1+xy) \leq \ln\left((1+x)(1+y)\right) = \ln(1+x) + \ln(1+y)$ with $x=e^{2^r}-1$ and $y=\norm[\bu]_1/\norm[\bu]_0$, \eqref{eqn:chapSparse-unknownByBphi-SRB-eachPeriod1-DeltaBound} yields
\begin{align*}
\Delta_r(\bu) & \leq 32 \, {y_T^*}^2 \norm[\bu]_0 2^r + 32 \, {y_T^*}^2 \norm[\bu]_0 \ln\left(1 +\frac{\norm[\bu]_1}{\norm[\bu]_0}\right) + 5 {y_T^*}^2 + 1~.
\end{align*}

Summing over $r = 0, \ldots, R$, we get
\begin{equation}
\label{eqn:chapSparse-unknownByBphi-SRB-eachPeriod3-sumDelta0}
\sum_{r=0}^R \Delta_r(\bu) \leq 32 \left(2^{R+1}-1\right) {y_T^*}^2 \norm[\bu]_0 + (R+1) \left(32 \, {y_T^*}^2 \norm[\bu]_0 \ln\left(1 +\frac{\norm[\bu]_1}{\norm[\bu]_0}\right) + 5 {y_T^*}^2 + 1\right)~.
\end{equation}

\noindent
\underline{First case: $R = 0$} \\
Substituting \eqref{eqn:chapSparse-unknownByBphi-SRB-eachPeriod3-sumDelta0} in \eqref{eqn:chapSparse-unknownByBphi-SRB-eachPeriod2}, we conclude the proof by noting that $A_T \geq 2 + \log_2 1 \geq 1$ and that $\ln\!\left(e + \sqrt{\sum_{t=1}^T \sum_{j=1}^d \phi_j^2(x_t)} \, \right) \geq 1$.\\

\noindent
\underline{Second case: $R \geq 1$} \\
Since $R \geq 1$, we have, by definition of $t_{R-1}$,
\begin{align*}
2^{R-1} & < \gamma_{t_{R-1}} \eqdef \ln\left(1 + \sqrt{\sum_{t=1}^{t_{R-1}} \sum_{j=1}^d \phi_j^2(x_t)} \, \right) \leq \ln\left(e + \sqrt{\sum_{t=1}^T \sum_{j=1}^d \phi_j^2(x_t)} \, \right)~.
\end{align*}
The last inequality entails that $2^{R+1}-1 \leq 4 \cdot 2^{R-1} \leq 4 \ln\!\left(e + \sqrt{\sum_{t=1}^T \sum_{j=1}^d \phi_j^2(x_t)} \, \right)$ and that $R+1 \leq 2 + \log_2 \ln \left(e + \sqrt{\sum_{t=1}^T \sum_{j=1}^d \phi_j^2(x_t)} \, \right) \eqdef A_T$. Therefore, one the one hand, via~(\ref{eqn:chapSparse-unknownByBphi-SRB-eachPeriod3-sumDelta0}),
\begin{align*}
\sum_{r=0}^R \Delta_r(\bu) & \leq 128 \, {y_T^*}^2 \norm[\bu]_0 \ln\left(e + \sqrt{\sum_{t=1}^T \sum_{j=1}^d \phi_j^2(x_t)} \, \right) + 32 \, {y_T^*}^2 A_T \norm[\bu]_0  \ln\left(1 + \frac{\norm[\bu]_1}{\norm[\bu]_0}\right) \label{eqn:chapSparse-unknownByBphi-SRB-eachPeriod3-sumDelta2} \\
& \quad + A_T \left(5 {y_T^*}^2 + 1\right)~,
\end{align*}
and, on the other hand,
\begin{equation*}
\label{eqn:chapSparse-unknownByBphi-SRB-eachPeriod3-RBound2}
4 (R+1) {y_T^*}^2 \leq 4 A_T \, {y_T^*}^2~.
\end{equation*}
Substituting the last two inequalities in  \eqref{eqn:chapSparse-unknownByBphi-SRB-eachPeriod2} and noting that ${y_T^*}^2 = \max_{1 \leq t \leq T} y_t^2$ concludes the proof.
\end{proofref}

\begin{proofref}{Corollary~\ref{cor:chapSparse-SRB-unknownByBphi}}
The proof is straightforward. In view of Theorem~\ref{thm:chapSparse-SRB-unknownByBphi}, we just need to check that the quantity (continuously extended in $s=0$)
\[
128 \Bigl(\max_{1 \leq t \leq T} y_t^2\Bigr) \, s \, \ln\left(e + \sqrt{\sum_{t=1}^T \sum_{j=1}^d \phi_j^2(x_t)} \, \right) + \, 32 \Bigl(\max_{1 \leq t \leq T} y_t^2\Bigr) A_T \, s \,  \ln\left(1 + \frac{U}{s}\right)
\]
is non-decreasing in $s \in \R_+$ and in $U \in \R_+$.

This is clear for $U$. The fact that it also non-decreasing in $s$ comes from the following remark. For all $U \geq 0$, the function $s \in (0,+\infty) \mapsto s \,  \ln(1+U/s)$ has a derivative equal to
\[
\ln\left(1 + \frac{U}{s}\right) - \frac{U/s}{1 + U/s} \qquad \textrm{for all} \quad s > 0~.
\]
From the elementary inequality
\[
\ln(1+u) = - \ln \left(\frac{1}{1+u}\right) \geq - \left(\frac{1}{1+u} - 1\right) = \frac{u}{1+u}~,
\]
which holds for all $u \in (-1,+\infty)$, the above derivative is nonnegative for all $s>0$ so that the continuous extension $s \in \R_+ \mapsto s \,  \ln\left(1 + U/s\right)$ is non-decreasing.
\end{proofref}

\subsection{Proofs of Theorem~\ref{thm:chapSparse-stochastic-tuningSimple} and Corollary~\ref{cor:chapSparse-stochastic-tuningSimple}}
\label{sec:chapSparse-proofs-random}

In this subsection, we set $\epsilon \eqdef Y - f(X)$, so that the pairs $(X_1,\epsilon_1),\ldots,(X_T,\epsilon_T)$ are independent copies of $(X,\epsilon) \in \cX \times \R$. We also define $\sigma \geq 0$ by
\[
\sigma^2 \eqdef \E\bigl[\epsilon^2\bigr] = \E\bigl[(Y - f(X))^2\bigr]~.
\]

\begin{proofref}{Theorem~\ref{thm:chapSparse-stochastic-tuningSimple}}
By Corollory~\ref{cor:chapSparse-SRB-unknownBy-tauSimple} and the definitions of $\widetilde{f}_t$ in~\eqref{eqn:chapSparse-stochastic-random-defAverage2} and $\widehat{y_t} \eqdef \widetilde{f}_t(X_t)$ in Figure~\ref{fig:chapSparse-continuouseEWA-varyingeta-def}, we have, \emph{almost surely},
\begin{align*}
\sum_{t=1}^T (Y_t - \widetilde{f}_t(X_t))^2 \leq & \inf_{\bu \in \R^d} \!\left\{\sum_{t=1}^T \bigl(Y_t - \bu \cdot \bphi(X_t)\bigr)^2 \!+ 32 \!\left(\max_{1 \leq t \leq T} Y_t^2\right) \norm[\bu]_0 \ln\!\left(1 \!+ \!\frac{\sqrt{d T} \norm[\bu]_1}{\norm[\bu]_0}\right) \!\right\} \\
& + \frac{1}{d T} \sum_{j=1}^d \sum_{t=1}^T \phi_j^2(X_t) + 5 \max_{1 \leq t \leq T} Y_t^2~.
\end{align*}
It remains to take the expectations of both sides with respect to $\bigl((X_1,Y_1), \ldots, (X_T,Y_T)\bigr)$. First note that for all $t=1, \ldots, T$, since $\epsilon_t \eqdef Y_t - f(X_t)$, we have
\begin{align*}
\E\left[\bigl(Y_t - \widetilde{f}_t(X_t)\bigr)^2\right] & = \E\left[\bigl(\epsilon_t + f(X_t) - \widetilde{f}_t(X_t)\bigr)^2\right] \\
& = \sigma^2 + \E\left[\bigl(f(X_t) - \widetilde{f}_t(X_t)\bigr)^2\right]~,
\end{align*}
since $\E\bigl[\epsilon_t^2\bigr] = \E\bigl[\epsilon^2\bigr] \eqdef \sigma^2$ on the one hand, and, on the other hand, $\widetilde{f}_t$ is a built on $(X_s,Y_s)_{1 \leq s \leq t-1}$ and $\E\bigl[\epsilon_t \big| (X_s,Y_s)_{1 \leq s \leq t-1}, X_t\bigr] = \E\bigl[\epsilon_t \big| X_t\bigr] = 0$ (from the independence of $(X_s,Y_s)_{1 \leq s \leq t-1}$ and $(X_t,Y_t)$ and by definition of $f$).

In the same way,
\begin{align*}
\E\left[\bigl(Y_t - \bu \cdot \bphi(X_t)\bigr)^2\right] & = \sigma^2 + \E\left[\bigl(f(X_t) - \bu \cdot \bphi(X_t)\bigr)^2\right]~.
\end{align*}
Therefore, by Jensen's inequality and the concavity of the infimum, the last inequality becomes, after taking the expectations of both sides,
\begin{align*}
T \sigma^2 + \sum_{t=1}^T \E\left[\bigl(f(X_t) - \widetilde{f}_t(X_t)\bigr)^2\right] \leq \inf_{\bu \in \R^d} & \Biggl\{T \sigma^2 + \sum_{t=1}^T \E\left[\bigl(f(X_t) - \bu \cdot \bphi(X_t)\bigr)^2\right] \\
& + 32 \, \E\left[\max_{1 \leq t \leq T} Y_t^2\right] \norm[\bu]_0 \, \ln\left(1 + \frac{\sqrt{d T} \, \norm[\bu]_1}{\norm[\bu]_0}\right)\Biggr\} \\
& + \frac{1}{d T} \sum_{j=1}^d \sum_{t=1}^T \E\left[\phi_j^2(X_t)\right] + 5 \, \E\left[\max_{1 \leq t \leq T} Y_t^2\right]~.
\end{align*}
Noting that the $T \sigma^2$ cancel out, dividing the two sides by $T$, and using the fact that $X_t \sim X$ in the right-hand side, we get
\begin{align*}
\frac{1}{T} \sum_{t=1}^T \E\left[\bigl(f(X_t)-\widetilde{f}_t(X_t)\bigr)^2\right] \leq \inf_{\bu \in \R^d} & \Biggl\{\norm[f-\bu \cdot \bphi]_{L^2}^2 \\
& + 32 \, \frac{\E\left[\max_{1 \leq t \leq T} Y_t^2\right]}{T} \norm[\bu]_0 \, \ln\left(1 + \frac{\sqrt{d T} \, \norm[\bu]_1}{\norm[\bu]_0}\right)\Biggr\} \\
& + \frac{1}{d T} \sum_{j=1}^d \norm[\phi_j]_{L^2}^2 + \, 5 \, \frac{\E\left[\max_{1 \leq t \leq T} Y_t^2\right]}{T}~.
\end{align*}
The right-hand side of the last inequality is exactly the upper bound stated in Theorem~\ref{thm:chapSparse-stochastic-tuningSimple}. To conclude the proof, we thus only need to check that $\E\bigl[\Arrowvert f-\widehat{f}_T \Arrowvert_{L^2}^2\bigr]$ is bounded from above by the left-hand side. But by definition of $\widehat{f}_T$ and by convexity of the square loss,
\begin{align*}
\E\!\left[\norm[f-\widehat{f}_T]_{L^2}^2\right] & \eqdef \E\left[\biggl(f(X)-\frac{1}{T} \sum_{t=1}^T \widetilde{f}_t(X)\biggr)^2\right] \\
& \leq \frac{1}{T} \sum_{t=1}^T \E\left[\bigl(f(X)-\widetilde{f}_t(X)\bigr)^2\right] = \frac{1}{T} \sum_{t=1}^T \E\left[\bigl(f(X_t)-\widetilde{f}_t(X_t)\bigr)^2\right]~.
\end{align*}
The last equality follows classically from the fact that, for all $t=1, \ldots, T$, $(X_s,Y_s)_{1 \leq s \leq t-1}$ (on which $\widetilde{f}_t$ is constructed) is independent from both $X_t$ and $X$ and the fact that $X_t \sim X$.
\end{proofref}

\begin{remark}
The fact that the inequality stated in Corollary~\ref{cor:chapSparse-SRB-unknownBy-tauSimple} has a leading constant equal to $1$ on individual sequences is crucial to derive in the stochastic setting an oracle inequality in terms of the (excess) risks $\E\!\left[\Arrowvert f-\widehat{f}_T \Arrowvert_{L^2}^2\right]$ and $\norm[f-\bu \cdot \bphi]_{L^2}^2$. Indeed, if the constant appearing in front of the infimum was equal to $C > 1$, then the $T \sigma^2$ would not cancel out in the previous proof, so that the resulting expected inequality would contain a non-vanishing additive term $(C-1) \sigma^2$.
\end{remark}

\begin{proofref}{Corollary~\ref{cor:chapSparse-stochastic-tuningSimple}}
We can apply Theorem~\ref{thm:chapSparse-stochastic-tuningSimple}. Then, to prove the upper bound on $\E\!\left[\Arrowvert f-\widehat{f}_T \Arrowvert_{L^2}^2\right]$, it suffices to show that
\begin{equation}
\frac{\E\left[\max_{1 \leq t \leq T} Y_t^2\right]}{T} \leq 2 \left(\frac{\E[Y]^2}{T} + \psi_T\right)~.
\label{eqn:chapSparse-stochastic-maxY2-upperbound}
\end{equation}
Recall that
$$
\psi_T \eqdef \frac{1}{T} \, \E\!\biggl[\max_{1 \leq t \leq T}\Bigl(Y_t - \E[Y_t]\Bigr)^2\biggr] = \frac{1}{T} \, \E\!\left[\max_{1 \leq t \leq T}(\Delta Y)_t^2\right]~,
$$
where we defined $(\Delta Y)_t \eqdef Y_t - \E[Y_t] = Y_t - \E[Y]$ for all $t=1, \ldots, T$.

From the elementary inequality $(x+y)^2 \leq 2 x^2 + 2 y^2$ for all $x,y \in \R$, we have
$$
\E\left[\max_{1 \leq t \leq T} Y_t^2\right] \eqdef \E\!\left[\max_{1 \leq t \leq T} \bigl(\E[Y] + (\Delta Y)_t\bigr)^2\right] \leq 2 \, \E[Y]^2 + 2 \, \E\left[\max_{1 \leq t \leq T} (\Delta Y)_t^2\right]~.
$$
Dividing both sides by $T$, we get (\ref{eqn:chapSparse-stochastic-maxY2-upperbound}).

As for the upper bound on $\psi_T$, since the $(\Delta Y)_t$, $1 \leq t \leq T$, are distributed as $\Delta Y$,  we can apply Lemmas~\ref{lem:chapSparse-maxIneq-SG}, \ref{lem:chapSparse-maxIneq-BEM}, and~\ref{lem:chapSparse-maxIneq-BM} in Appendix~\ref{apx:chapSparse-maxIneq} to bound $\psi_T$ from above under the assumptions $\bigl(\rm{SG}(\sigma^2)\bigr)$, $\bigl(\rm{BEM}(\alpha,M)\bigr)$, and $\bigl(\rm{BM}(\alpha,M)\bigr)$ respectively (the upper bound under $\bigl(\rm{BD}(B)\bigr)$ is straightforward):
\[
\E\!\left[\max_{1 \leq t \leq T} (\Delta Y)_t^2\right] \leq
\begin{cases}
B^2 & \mbox{if Assumption } \bigl(\rm{BD}(B)\bigr) \mbox{holds, } \\
\sigma^2 + 2 \sigma^2 \ln(2 e T) & \mbox{if Assumption } \bigl(\rm{SG}(\sigma^2)\bigr) \mbox{holds, } \\
\frac{\ln^2\bigl((M+e) T\bigr)}{\alpha^2} & \mbox{if Assumption } \bigl(\rm{BEM}(\alpha,M)\bigr) \mbox{holds, } \\
(M T)^{2/\alpha} & \mbox{if Assumption } \bigl(\rm{BM}(\alpha,M)\bigr) \mbox{holds~.}
\end{cases}
\]
\end{proofref}

\begin{remark}
\label{rmk:chapSparse-stochastic-EYnotNecessary}
If $T \geq 2$, then the two ``bias'' terms $\E[Y]^2 / T$ appearing in Corollary~\ref{cor:chapSparse-stochastic-tuningSimple} can be avoided, at least at the price of a multiplicative factor of $2T/(T-1)\leq 4$. It suffices to use a slightly more sophisticated online clipping defined as follows. The first round $t=1$ is only used to observe $Y_1$. Then, the algorithm $\rm{SeqSEW}^*_{\tau}$ is run with $\tau = 1/\sqrt{d T}$ from round $2$ up to round $T$ with the following important modification: instead of truncating the predictions to $[-B_t,B_t]$, which is best suited to the case $\E[Y]=0$, we truncate them to the interval
$$
\bigl[Y_1-B'_t,Y_1+B'_t\bigr]~, \quad \textrm{where} \quad B'_t \eqdef \max_{1 \leq s \leq t-1} |Y_s-Y_1|~.
$$
If $\eta_t$ is changed accordingly, that is, if $\eta_t = 1 / (8 {B'_t}^2)$, then it easy to see that the resulting procedure $\widehat{f}_T \eqdef \frac{1}{T-1} \sum_{s=2}^T \tilde{f}_s$ (where $\tilde{f}_2,\ldots,\tilde{f}_T$ are the estimators output by $\rm{SeqSEW}^*_{\tau}$) satisfies
\begin{align*}
\nonumber
\E\!\left[\norm[f-\widehat{f}_T]_{L^2}^2\!\right] \leq & \inf_{\bu \in \R^d} \!\left\{\norm[f-\bu \cdot \bphi]_{L^2}^2 + 64 \!\left( \frac{\Var [Y]}{T-1} + \psi_{T-1} \right) \norm[\bu]_0 \ln\!\left(1 + \frac{\sqrt{d T} \norm[\bu]_1}{\norm[\bu]_0}\right)\!\right\}  \\
\nonumber
& + \frac{1}{d T} \sum_{j=1}^d \norm[\phi_j]_{L^2}^2 + 10 \!\left( \frac{\Var [Y]}{T-1} + \psi_{T-1} \right)~,
\end{align*}
where $\Var [Y] \eqdef \E\bigl[(Y-\E[Y])^2\bigr]$. Comparing the last bound to that of Corollary~\ref{cor:chapSparse-stochastic-tuningSimple}, we note that the two terms $\E[Y]^2 / T$ are absent, and that we loose a multiplicative factor at most of~$4$ since $\Var [Y] \leq \E\bigl[\max_{2 \leq t \leq T} (Y_t-\E[Y_t])^2\bigr] \eqdef (T-1) \psi_{T-1}$ so that
$$
\frac{\Var [Y]}{T-1} + \psi_{T-1} \leq 2 \psi_{T-1} \leq 2 \left(\frac{T}{T-1}\right) \psi_{T} \leq 4 \psi_T~.
$$
\end{remark}

\begin{remark}
\label{rmk:chapSparse-stochastic-assump-decoupling}
We mentioned after Corollary~\ref{cor:chapSparse-stochastic-tuningSimple} that each of the four assumptions on $\Delta Y$ is fulfilled as soon as both the distribution of $f(X)-\E[f(X)]$ and the conditional distribution of $\epsilon$ (conditionally on $X$) satisfy the same type of assumption. It actually extends to the more general case when the conditional distribution of $\epsilon$ given $X$ is replaced with the distribution of $\epsilon$ itself (without conditioning). This relies on the elementary upper bound
\begin{align*}
\nonumber
\E\!\left[\max_{1 \leq t \leq T} (\Delta Y)_t^2\right] & = \E\!\left[\max_{1 \leq t \leq T} \bigl(f(X_t)-\E[f(X)] + \epsilon_t\bigr)^2\right] \\
\nonumber
& \leq 2 \, \E\!\left[\max_{1 \leq t \leq T} \bigl(f(X_t)-\E[f(X)]\bigr)^2\right] + 2 \, \E\!\left[\max_{1 \leq t \leq T} \epsilon_t^2\right]~.
\end{align*}
From the last inequality, we can also see that assumptions of different nature can be made on $f(X)-\E[f(X)]$ and $\epsilon$, such as the assumptions given in~\eqref{rmk:chapSparse-stochastic-assump1} or in~\eqref{rmk:chapSparse-stochastic-assump2}.
\end{remark}

\subsection{Proofs of Theorem~\ref{thm:chapSparse-stochastic-fixedDesign-tuningSimple} and Corollary~\ref{cor:chapSparse-stochastic-fixedDesign-tuningSimple}}
\label{sec:chapSparse-proofs-fixed}

\begin{proofref}{Theorem~\ref{thm:chapSparse-stochastic-fixedDesign-tuningSimple}}
The proof follows the sames lines as in the proof of Theorem~\ref{thm:chapSparse-stochastic-tuningSimple}. We thus only sketch the main arguments. In the sequel, we set $\sigma^2 \eqdef \E\bigl[\epsilon_1^2]$.

Applying Corollory~\ref{cor:chapSparse-SRB-unknownBy-tauSimple} we have, \emph{almost surely},
\begin{align*}
\nonumber
\sum_{t=1}^T \bigl(Y_t - \widetilde{f}_t(x_t)\bigr)^2 \leq & \inf_{\bu \in \R^d} \!\left\{\sum_{t=1}^T \bigl(Y_t - \bu \cdot \bphi(x_t)\bigr)^2 + 32 \!\left(\max_{1 \leq t \leq T} Y_t^2\right) \norm[\bu]_0 \ln\!\left(1 + \frac{\sqrt{d T} \norm[\bu]_1}{\norm[\bu]_0}\right)\!\right\} \\
\nonumber
& + \frac{1}{d T} \sum_{j=1}^d \sum_{t=1}^T \phi_j^2(x_t) + 5 \, \max_{1 \leq t \leq T} Y_t^2~.
\end{align*}
Taking the expectations of both sides, expanding the squares $\bigl(Y_t - \widetilde{f}_t(x_t)\bigr)^2$ and $\bigl(Y_t - \bu \cdot \bphi(x_t)\bigr)^2$, noting that two terms $T \sigma^2$ cancel out,\footnote{Note that $\E\bigl[ (f(x_t)-\tilde{f}(x_t)) \epsilon_t \bigr] = 0$ since $\tilde{f}_t(x_t)$ and $\epsilon_t$ are independent. This is due to the fact that $\tilde{f}_t$ is built from the past data only. In particular, truncating the predictions to $B=\max_{1 \leq t \leq T} |Y_t|$ might not work. A similar comment could be made in the random design case (Section~\ref{sec:chapSparse-adaptivityVariance-random}).} and then dividing both sides by $T$, we get
\begin{align*}
\nonumber
\E\!\left[\frac{1}{T} \sum_{t=1}^T \bigl(f(x_t) - \widetilde{f}_t(x_t)\bigr)^2\right] \leq & \inf_{\bu \in \R^d} \Biggl\{\frac{1}{T} \sum_{t=1}^T \bigl(f(x_t) - \bu \cdot \bphi(x_t)\bigr)^2 \\
\nonumber
& + 32 \, \frac{\E\!\left[\max_{1 \leq t \leq T} Y_t^2\right]}{T} \norm[\bu]_0 \, \ln\left(1 + \frac{\sqrt{d T} \, \norm[\bu]_1}{\norm[\bu]_0}\right)\Biggr\} \\
\nonumber
& + \frac{1}{d T^2} \sum_{j=1}^d \sum_{t=1}^T \phi_j^2(x_t) + 5 \, \frac{\E\!\left[\max_{1 \leq t \leq T} Y_t^2\right]}{T}~.
\end{align*}
The right-hand side is exactly the upper bound stated in Theorem~\ref{thm:chapSparse-stochastic-fixedDesign-tuningSimple}. We thus only need to check that
\begin{equation}
\E\!\left[\frac{1}{T} \sum_{t=1}^T \bigl(f(x_t)-\widehat{f}_T(x_t)\bigr)^2\right] \leq \E\!\left[\frac{1}{T} \sum_{t=1}^T \bigl(f(x_t) - \widetilde{f}_t(x_t)\bigr)^2\right]~.
\label{eqn:chapSparse-stochastic-fixedDesign-Jensen}
\end{equation}
This is an equality if the $x_t$ are all distinct. In general we get an inequality which follows from the convexity of the square loss. Indeed, by definition of $n_x$, we have, almost surely,
\begin{align*}
\sum_{t=1}^T \bigl(f(x_t)-\widehat{f}_T(x_t)\bigr)^2 & = \sum_{x \in \{x_1, \ldots, x_T\}} \, \sum_{\substack{1 \leq t \leq T\\ t:x_t=x}} \bigl(f(x_t)-\widehat{f}_T(x_t)\bigr)^2 = \sum_{x \in \{x_1, \ldots, x_T\}} n_x \, \bigl(f(x)-\widehat{f}_T(x)\bigr)^2 \\
& = \sum_{x \in \{x_1, \ldots, x_T\}} n_x \, \Biggl(f(x)-\frac{1}{n_x} \sum_{\substack{1 \leq t \leq T\\ t:x_t=x}} \widetilde{f}_t(x)\Biggr)^2 \\
& \leq \sum_{x \in \{x_1, \ldots, x_T\}} n_x \, \frac{1}{n_x} \sum_{\substack{1 \leq t \leq T\\ t:x_t=x}} \bigl(f(x)-\widetilde{f}_t(x)\bigr)^2 = \sum_{t=1}^T \bigl(f(x_t) - \widetilde{f}_t(x_t)\bigr)^2~,
\end{align*}
where the second line is by definition of $\widehat{f}_T$ and where the last line follows from Jensen's inequality. Dividing both sides by $T$ and taking their expectations, we get (\ref{eqn:chapSparse-stochastic-fixedDesign-Jensen}), which concludes the proof.
\end{proofref}

\begin{proofref}{Corollary~\ref{cor:chapSparse-stochastic-fixedDesign-tuningSimple}}
First note that
\begin{align*}
\E\!\left[\max_{1 \leq t \leq T} Y_t^2\right] & \eqdef \E\!\left[\max_{1 \leq t \leq T} \bigl(f(x_t) + \epsilon_t\bigr)^2\right] \leq 2 \left(\max_{1 \leq t \leq T} f^2(x_t) + \E\Bigl[\max_{1 \leq t \leq T} \epsilon_t^2\Bigr]\right)~.
\end{align*}
The proof then follows exactly the same lines as for Corollary~\ref{cor:chapSparse-stochastic-tuningSimple} with the sequence $(\epsilon_t)$ instead of the sequence $\bigl((\Delta Y)_t\bigr)$.
\end{proofref}

\section{Tools}
\label{sec:chapSparse-tools}

Next we provide several (in)equalities that prove to be useful throughout the paper.

\subsection{A Duality Formula for the Kullback-Leibler Divergence}
\label{apx:dualityKL}

We recall below a key duality formula satisfied by the Kullback-Leibler divergence and whose proof can be found, for example, in the monograph by \citet[{pp.}\ 159--160]{Catoni01StFlour}. We use the notations of Section~\ref{sec:chapSparse-setting}.

\begin{proposition}
\label{prop:appendix-dualityKL}
For any measurable space $(\Theta, \mathcal{B})$, any probability distribution $\pi$ on $(\Theta, \mathcal{B})$, and any measurable function $h: \Theta \rightarrow [a,+\infty)$ bounded from below (by some $a \in \R$), we have
\[
- \ln \int_{\Theta} e^{-h} \dd\pi = \inf_{\rho \in \cM_1^+(\Theta)} \left\{\int_{\Theta} h \, \dd\rho \, + \, \KL(\rho,\pi)\right\}~,
\]
where $\cM_1^+(\Theta)$ denotes the set of all probability distributions on $(\Theta, \mathcal{B})$, and where the expectations $\int_{\Theta} h \, \dd\rho \in [a,+\infty]$ are always well defined since $h$ is bounded from below.
\end{proposition}

\subsection{Some Tools to Exploit Our PAC-Bayesian Inequalities}
\label{apx:chapSparse-tools-PACB}
In this subsection we recall two results needed for the derivation of Proposition~\ref{prop:chapSparse-SRB-known} and Proposition~\ref{prop:chapSparse-SRB-unknownBy} from the PAC-Bayesian inequalities (\ref{eqn:chapSparse-PACB-known-notTruncated}) and (\ref{eqn:chapSparse-PACB-unknownBy-notTruncated}). The proofs are due to \citet{DaTsy07SEW, DaTsy08SEW} and we only reproduce them for the convenience of the reader.\footnote{The notations are however slightly modified because of the change in the statistical setting and goal. The target predictions $(f(x_1), \ldots, f(x_T))$ are indeed replaced with the observations $(y_1, \ldots, y_T)$ and the prediction loss $\Arrowvert f-f_{\bu}\Arrowvert_n^2$ is replaced with the cumulative loss $\sum_{t=1}^T \bigl(y_t - \bu \cdot \bphi(x_t)\bigr)^2$. Moreover, the analysis of the present proof is slightly simpler since we just need to consider the case $L_0 = +\infty$ according to the notations of Theorem~5 by \citet{DaTsy08SEW}.}

For any $\bu^* \in  \R^d$ and $\tau > 0$, define $\rho_{\bu^*,\tau}$ as the translated of $\pi_{\tau}$ at $\bu^*$, namely,
\begin{equation}
\rho_{\bu^*,\tau} \eqdef \frac{\d\pi_{\tau}}{\d\bu} (\bu - \bu^*) \, \d\bu = \prod_{j=1}^d \frac{(3/\tau) \, \d u_j}{2 \bigl(1+|u_j - u^*_j|/\tau\bigr)^4}~.
\label{eqn:chapSparse-PACBupperbound-defrho}
\end{equation}

\begin{lemma}
\label{lem:chapSparse-PACBupperbound-int}
For all $\bu^* \in  \R^d$ and $\tau > 0$, the probability distribution $\rho_{\bu^*,\tau}$ satisfies
\[
\int_{\R^d} \sum_{t=1}^T \bigl(y_t - \bu \cdot \bphi(x_t)\bigr)^2 \rho_{\bu^*,\tau}(\dd\bu) = \sum_{t=1}^T \bigl(y_t - \bu^* \cdot \bphi(x_t)\bigr)^2 + \tau^2 \sum_{j=1}^d \sum_{t=1}^T \phi_j^2(x_t)~.
\]
\end{lemma}

\begin{lemma}
\label{lem:chapSparse-PACBupperbound-KL}
For all $\bu^* \in  \R^d$ and $\tau > 0$, the probability distribution $\rho_{\bu^*,\tau}$ satisfies
\[
\KL(\rho_{\bu^*,\tau}, \pi_{\tau}) \leq 4 \norm[\bu^*]_0 \ln\left(1 + \frac{\norm[\bu^*]_1}{\norm[\bu^*]_0 \tau}\right)~.
\]
\end{lemma}

\begin{proofref}{Lemma~\ref{lem:chapSparse-PACBupperbound-int}}
For all $t \in \{1, \ldots, T\}$ we expand the square $\bigl(y_t - \bu \cdot \bphi(x_t)\bigr)^2 = \bigl(y_t - \bu^* \cdot \bphi(x_t) + (\bu^* - \bu) \cdot \bphi(x_t) \bigr)^2$  and use the linearity of the integral to get
\begin{align}
& \int_{\R^d} \sum_{t=1}^T \bigl(y_t - \bu \cdot \bphi(x_t)\bigr)^2 \rho_{\bu^*,\tau}(\d\bu) \label{eqn:chapSparse-PACBupperbound-squareExpand-final} \\
& \qquad = \sum_{t=1}^T \bigl(y_t - \bu^* \cdot \bphi(x_t)\bigr)^2 + \sum_{t=1}^T \int_{\R^d} \bigl((\bu^* - \bu) \cdot \bphi(x_t)\bigr)^2 \rho_{\bu^*,\tau}(\d\bu) \nonumber \\
& \qquad \quad + \underbrace{\sum_{t=1}^T 2 \bigl(y_t - \bu^* \cdot \bphi(x_t)\bigr) \int_{\R^d}(\bu^* - \bu) \cdot \bphi(x_t) \, \rho_{\bu^*,\tau}(\d\bu)}_{= 0} \nonumber
\end{align}
The last sum equals zero by symmetry of $\rho_{\bu^*,\tau}$ around $\bu^*$, which yields $\displaystyle{\int_{\R} \bu \, \rho_{\bu^*,\tau}(\d\bu)= \bu^*}$. As for the second sum of the right-hand side, it can be bounded from above similarly. Indeed, expanding the inner product and then the square $\bigl((\bu^* - \bu) \cdot \bphi(x_t)\bigr)^2$ we have, for all  $t=1,\ldots,T$,
\[
\bigl((\bu^* - \bu) \cdot \bphi(x_t)\bigr)^2 = \sum_{j=1}^d (u^*_j - u_j)^2 \phi_j^2(x_t) + \sum_{1 \leq j \neq k \leq d} (u^*_j - u_j) (u^*_k - u_k) \, \phi_j(x_t) \, \phi_k(x_t)~.
\]
By symmetry of $\rho_{\bu^*,\tau}$ around $\bu^*$ and the fact that $\rho_{\bu^*,\tau}$ is a product-distribution, we get
\begin{align}
\sum_{t=1}^T \int_{\R^d} \bigl((\bu^* - \bu) \cdot \bphi(x_t)\bigr)^2 \rho_{\bu^*,\tau}(\d\bu) = & \sum_{t=1}^T \sum_{j=1}^d \phi_j^2(x_t) \int_{\R^d} (u^*_j - u_j)^2 \rho_{\bu^*,\tau}(\d\bu) \, + 0 \nonumber \\
= & \sum_{t=1}^T \sum_{j=1}^d \phi_j^2(x_t) \int_{\R} (u^*_j - u_j)^2 \frac{(3/\tau) \, \d u_j}{2 \bigl(1+|u_j-u^*_j|/\tau\bigr)^4} \label{eqn:chapSparse-PACBupperbound-tau2-1} \\[-0.3cm]
= & \tau^2 \sum_{t=1}^T \sum_{j=1}^d \phi_j^2(x_t) \int_{\R} \frac{3 t^2 \d t}{2 (1+|t|)^4} \label{eqn:chapSparse-PACBupperbound-tau2-2} \\[-0.1cm]
= & \tau^2 \sum_{t=1}^T \sum_{j=1}^d \phi_j^2(x_t)~, \label{eqn:chapSparse-PACBupperbound-tau2-3}
\end{align}
where \eqref{eqn:chapSparse-PACBupperbound-tau2-1} follows by definition of $\rho_{\bu^*,\tau}$, where \eqref{eqn:chapSparse-PACBupperbound-tau2-2} is obtained by the change of variables $t = (u_j-u^*_j) / \tau$, and where \eqref{eqn:chapSparse-PACBupperbound-tau2-3} follows from the equality $\displaystyle{\int_{\R} \frac{3 t^2 \d t}{2 \bigl(1+|t|\bigr)^4} = 1}$ that can be proved by integrating by parts. Substituting \eqref{eqn:chapSparse-PACBupperbound-tau2-3} into \eqref{eqn:chapSparse-PACBupperbound-squareExpand-final} concludes the proof.
\end{proofref}

\begin{proofref}{Lemma~\ref{lem:chapSparse-PACBupperbound-KL}}
By definition of $\rho_{\bu^*,\tau}$ and $\pi_{\tau}$, we have
\begin{align}
\KL(\rho_{\bu^*,\tau}, \pi_{\tau}) \eqdef & \int_{\R^d} \left(\ln \frac{\d \rho_{\bu^*,\tau}}{\d \pi_{\tau}}(\bu)\right) \rho_{\bu^*,\tau}(\d\bu) = \int_{\R^d} \left(\ln \prod_{j=1}^d \frac{\bigl(1+|u_j|/\tau\bigr)^4}{\bigl(1+|u_j-u^*_j|/\tau\bigr)^4} \right) \rho_{\bu^*,\tau}(\d\bu) \nonumber \\
= & \, 4 \int_{\R^d} \biggl(\sum_{j=1}^d \ln \frac{1+|u_j|/\tau}{1+|u_j-u^*_j|/\tau}\biggr) \, \rho_{\bu^*,\tau}(\d\bu)~. \label{eqn:chapSparse-PACBupperbound-KL-expand}
\end{align}
But, for all $\bu \in \R^d$, by the triangle inequality,
\[
1 + |u_j|/\tau \leq 1 + |u^*_j|/\tau + |u_j - u^*_j|/\tau \leq \bigl(1 + |u^*_j|/\tau\bigr) \bigl(1 + |u_j - u^*_j|/\tau\bigr)~,
\]
so that Equation~(\ref{eqn:chapSparse-PACBupperbound-KL-expand}) yields the upper bound
\begin{align*}
\KL(\rho_{\bu^*,\tau}, \pi_{\tau}) & \leq 4 \, \sum_{j=1}^d \ln\left(1+|u^*_j|/\tau\right) = 4 \, \sum_{j:u^*_j \neq 0} \ln\left(1+|u^*_j|/\tau\right)~.
\end{align*}
We now recall that $\norm[\bu^*]_0\eqdef \bigl|\{j: u^*_j \neq 0\}\bigr|$ and apply Jensen's inequality to the concave function $x \in (-1,+\infty) \longmapsto \ln(1+x)$ to get
\begin{align*}
\sum_{j:u^*_j \neq 0} \ln\left(1+|u^*_j|/\tau\right) & =  \norm[\bu^*]_0 \, \frac{1}{\norm[\bu^*]_0} \sum_{j:u^*_j \neq 0} \ln\left(1+|u^*_j|/\tau\right) \leq \norm[\bu^*]_0 \ln\left(1 + \frac{\sum_{j:u^*_j \neq 0} |u^*_j|}{\norm[\bu^*]_0 \tau}\right) \\
& \leq \norm[\bu^*]_0 \ln\left(1 + \frac{\norm[\bu^*]_1}{\norm[\bu^*]_0 \tau}\right)~.
\end{align*}
This concludes the proof.
\end{proofref}

\subsection{Some Maximal Inequalities}
\label{apx:chapSparse-maxIneq}

Next we prove three maximal inequalities needed for the derivation of Corollaries~\ref{cor:chapSparse-stochastic-tuningSimple} and~\ref{cor:chapSparse-stochastic-fixedDesign-tuningSimple} from Theorems~\ref{thm:chapSparse-stochastic-tuningSimple} and~\ref{thm:chapSparse-stochastic-fixedDesign-tuningSimple} respectively. Their proofs are quite standard but we provide them for the convenience of the reader.

\begin{lemma}
\label{lem:chapSparse-maxIneq-SG}
Let $Z_1, \ldots, Z_T$ be $T \geq 1$ (centered) real random variables such that, for a given constant $\nu \geq 0$, we have
\begin{equation}
\label{eqn:chapSparse-maxIneq-SG-assum}
\forall t \in \{1, \ldots, T\}, \quad \forall \lambda \in \R, \quad \E\!\left[e^{\lambda Z_t}\right] \leq e^{\lambda^2 \nu / 2}~.
\end{equation}
Then,
\[
\E\!\left[\max_{1 \leq t \leq T} Z_t^2\right] \leq 2 \nu \ln(2 \mathrm{e} T)~.
\]
\end{lemma}

\begin{lemma}
\label{lem:chapSparse-maxIneq-BEM}
Let $Z_1, \ldots, Z_T$ be $T \geq 1$ real random variables such that, for some given constants $\alpha > 0$ and $M > 0$, we have
\[
\forall t \in \{1, \ldots, T\}, \quad \E\!\left[e^{\alpha |Z_t|}\right] \leq M~.
\]
Then,
\[
\E\!\left[\max_{1 \leq t \leq T} Z_t^2\right] \leq \frac{\ln^2\bigl((M+e)T\bigr)}{\alpha^2}~.
\]
\end{lemma}

\begin{lemma}
\label{lem:chapSparse-maxIneq-BM}
Let $Z_1, \ldots, Z_T$ be $T \geq 1$ real random variables such that, for some given constants $\alpha > 2$ and $M > 0$, we have
\[
\forall t \in \{1, \ldots, T\}, \quad \E\bigl[|Z_t|^{\alpha}\bigr] \leq M~.
\]
Then,
\[
\E\!\left[\max_{1 \leq t \leq T} Z_t^2\right] \leq (M T)^{2 / \alpha}~.
\]
\end{lemma}

\begin{proofref}{Lemma~\ref{lem:chapSparse-maxIneq-SG}}
Let $t \in \{1, \ldots, T\}$. From the subgaussian assumption~(\ref{eqn:chapSparse-maxIneq-SG-assum})  it is well known (see, e.g., \citealt[Chapter~2]{Massart03StFlour}) that for all $x \geq 0$, we have
\[
\forall t \in \{1,\ldots,T\}~, \quad \Prob\bigl(|Z_t| > x\bigr) \leq 2 e^{-x^2/(2 \nu)}~.
\]
Let $\delta \in (0,1)$. By the change of variables $x = \sqrt{2 \nu \ln (2 T/\delta)}$, the last inequality entails that, for all $t=1,\ldots,T$, we have $|Z_t| \leq \sqrt{2 \nu \ln (2T/\delta)}$ with probability at least $1-\delta/T$. Therefore, by a union bound, we get, with probability at least $1-\delta$,
\[
\forall t \in \{1, \ldots, T\}~, \quad |Z_t| \leq \sqrt{2 \nu \ln (2T/\delta)}~.
\]
As a consequence, with probability at least $1-\delta$,
\[
\max_{1 \leq t \leq T} Z_t^2 \leq 2 \nu \ln (2T/\delta) \leq 2 \nu \ln (1/\delta) + 2 \nu \ln (2T)~.
\]
It now just remains to integrate the last inequality over $\delta \in (0,1)$ as is made precise below. By the change of variables $\delta = e^{-z}$, the latter inequality yields
\begin{equation}
\label{eqn:chapSparse-maxIneq-SG-beforeInt}
\forall z > 0~, \quad \Prob\Biggl[\left(\frac{\max_{1 \leq t \leq T} Z_t^2 - 2 \nu \ln (2T) }{2 \nu}\right)_+ > z \Biggr] \leq e^{-z}~,
\end{equation}
where for all $x \in \R$, $x_+ \eqdef \max\{x,0\}$ denotes the positive part of $x$. Using the well-known fact that $\E[\xi]=\int_0^{+\infty} \Prob(\xi>z) \dd z$ for all nonnegative real random variable $\xi$, we get
\begin{align*}
\E\Biggl[\frac{\max_{1 \leq t \leq T} Z_t^2 - 2 \nu \ln (2T) }{2 \nu}\Biggr] & \leq \E\Biggl[\left(\frac{\max_{1 \leq t \leq T} Z_t^2 - 2 \nu \ln (2T) }{2 \nu}\right)_+\Biggr] \\
& = \int_0^{+\infty} \Prob\Biggl[\left(\frac{\max_{1 \leq t \leq T} Z_t^2 - 2 \nu \ln (2T) }{2 \nu}\right)_+ > z \Biggr] \dd z \\
& \leq \int_0^{+\infty} e^{-z} \dd z~ = 1~,
\end{align*}
where the last line follows from~(\ref{eqn:chapSparse-maxIneq-SG-beforeInt}) above. Rearranging terms, we get $\E\!\left[\max_{1 \leq t \leq T} Z_t^2\right] \leq 2 \nu + 2 \nu \ln(2T)$, which concludes the proof.
\end{proofref}

\begin{proofref}{Lemma~\ref{lem:chapSparse-maxIneq-BEM}}
We first need the following definitions. Let $\psi_{\alpha}: \R_+ \to \R$ be a convex majorant of $x \mapsto e^{\alpha \sqrt{x}}$ on $\R_+$ defined by
\begin{numcases}{\psi_{\alpha}(x) \eqdef}
\nonumber
e & if $x < 1/\alpha^2$~, \\
\nonumber
e^{\alpha \sqrt{x}} & if $x \geq 1/\alpha^2$~.
\end{numcases}
We associate with $\psi_{\alpha}$ its generalized inverse $\psi_{\alpha}^{-1}: \R \to \R_+$ defined by
\begin{numcases}{\psi_{\alpha}^{-1}(y) =}
\nonumber
1/\alpha^2 & if $y < e$~, \\
\nonumber
(\ln y)^2 / \alpha^2 & if $y \geq e$~.
\end{numcases}
Elementary manipulations show that:
\begin{itemize}
\item $\psi_{\alpha}$ is nondecreasing and convex on $\R_+$;
\item $\psi_{\alpha}^{-1}$ is nondecreasing on $\R$;
\item $x \leq \psi_{\alpha}^{-1}\bigl(\psi_{\alpha}(x)\bigr)$ for all $x \in \R_+$.
\end{itemize}

The proof is based on a Pisier-type argument as is done, for example, by \citet[Lemma~2.3]{Massart03StFlour} to prove the maximal inequality $\E\!\left[\max_{1 \leq t \leq T} \xi_t \right] \leq \sqrt{2 \nu \ln T}$ for all subgaussian real random variables $\xi_t$, $1 \leq t \leq T$, with common variance factor $\nu \geq 0$.

From the inequality $x \leq \psi_{\alpha}^{-1}\bigl(\psi_{\alpha}(x)\bigr)$ for all $x \in \R_+$ we have
\begin{align*}
\E\left[\max_{1 \leq t \leq T} Z_t^2\right] & \leq \psi_{\alpha}^{-1}\left(\psi_{\alpha}\left(\E\Bigl[\max_{1 \leq t \leq T} Z_t^2\Bigr]\right)\right) \\
& \leq \psi_{\alpha}^{-1}\left(\E\left[\psi_{\alpha}\Bigl(\max_{1 \leq t \leq T} Z_t^2\Bigr)\right]\right) = \psi_{\alpha}^{-1}\left(\E\left[\max_{1 \leq t \leq T} \psi_{\alpha}\bigl(Z_t^2\bigr)\right]\right)~,
\end{align*}
where the last two inequalities follow by Jensen's inequality (since $\psi_{\alpha}$ is convex) and the fact that both $\psi_{\alpha}^{-1}$ and $\psi_{\alpha}$ are nondecreasing.

Since $\psi_{\alpha} \geq 0$ and $\psi_{\alpha}^{-1}$ is nondecreasing we get
\begin{align*}
\E\left[\max_{1 \leq t \leq T} Z_t^2\right] & \leq \psi_{\alpha}^{-1}\left(\E\left[\sum_{t=1}^T \psi_{\alpha}\bigl(Z_t^2\bigr)\right]\right) = \psi_{\alpha}^{-1}\left(\sum_{t=1}^T \E\Bigl[ \psi_{\alpha}\bigl(Z_t^2\bigr)\Bigr]\right) \\
& \leq \psi_{\alpha}^{-1}\left(\sum_{t=1}^T \E\Bigl[e^{\alpha |Z_t|} + e\Bigr]\right) \\
& \leq \psi_{\alpha}^{-1}\bigl(M T + e T\bigr) = \frac{\ln^2 \bigl(M T + e T\bigr)}{\alpha^2}~,
\end{align*}
where the second line follows from the inequality $\psi_{\alpha}(x) \leq e + e^{\alpha \sqrt{x}}$ for all $x \in \R_+$, and where the last line follows from the bounded exponential moment assumption and the definition of $\psi_{\alpha}^{-1}$. It concludes the proof.
\end{proofref}

\begin{proofref}{Lemma~\ref{lem:chapSparse-maxIneq-BM}}
As in the previous proof, we have, by Jensen's inequality and the fact that $x \mapsto x^{\alpha / 2}$ is convex and nondecreasing on $\R_+ $ (since $\alpha > 2$),
\begin{align*}
\E\left[\max_{1 \leq t \leq T} Z_t^2\right] & \leq \E\left[\left(\max_{1 \leq t \leq T} Z_t^2\right)^{\alpha / 2}\right]^{2 / \alpha} = \E\left[\max_{1 \leq t \leq T} \big|Z_t\big|^{\alpha}\right]^{2 / \alpha} \\
& \leq \E\left[\sum_{t=1}^T \big|Z_t\big|^{\alpha}\right]^{2 / \alpha} \leq (M T)^{2 / \alpha}
\end{align*}
by the bounded-moment assumption, which concludes the proof.
\end{proofref}

\bibliography{journal,reference}

\end{document}